\pdfoutput=1

\documentclass[11pt]{article}

\usepackage{acl}
\usepackage{times}
\usepackage{latexsym}
\usepackage[T1]{fontenc}
\usepackage[utf8]{inputenc}
\usepackage{microtype}
\usepackage{amsmath}
\usepackage{amssymb}
\usepackage{booktabs}
\usepackage{graphicx}
\usepackage{xspace}
\usepackage{algorithm}
\usepackage{algpseudocode}
\usepackage{enumitem}
\usepackage{xcolor}
\usepackage{listings}
\usepackage{tikz}
\usepackage{inconsolata}    

\newcommand{\system}{\textsc{SIRIUS-SQL}\xspace}
\newcommand{\bird}{\textsc{BIRD}\xspace}
\newcommand{\spider}{\textsc{Spider}\xspace}

\definecolor{deltagreen}{HTML}{2E7D32}
\definecolor{deltared}{HTML}{C62828}
\DeclareRobustCommand*\circled[1]{\tikz[baseline=(char.base)]{%
  \node[shape=circle,draw,inner sep=1pt,font=\scriptsize\bfseries] (char) {#1};}}

\newcommand{\drop}[1]{\,{\scriptsize\textcolor{deltared}{$\downarrow$#1}}}

\title{\system: Anchoring Multi-Candidate Text-to-SQL in Execution Feedback}

\author{
      Leo Luo{$^{1}$} , Haining Xie{$^{1}$} , Siqi Shen{$^{1,2}$} , Zhipeng Ma{$^{1}$} , Rui Ling{$^{1,2}$} \\
      \textbf{Hang Xu{$^{1}$} , Hefeng Jiang{$^{1}$} , Dingwei Chen{$^{1}$} , Yang Li{$^{1}\thanks{~~Corresponding author.}$} , Peng Chen{$^{1}$} ,
  Jie Jiang{$^{1}$}} \\
      $^1$TEG, Tencent Inc., China \\
      $^2$Peking University, China \\
      \texttt{
      leoluo00@outlook.com} \\
      \texttt{
      \{hainingxie,siqishen,allenzpma,ruiling,hhangxu\}@tencent.com} \\
      \texttt{
      \{hefengjiang,espadachen,thomasyngli,pengchen,zeus\}@tencent.com}
  }

\begin{document}
\maketitle

\begin{abstract}
Text-to-SQL on complex schemas is unreliable on a single pass, so recent systems generate multiple SQL candidates and let voting filter out errors.
Yet voting alone is not enough, because the multi-candidate recipe has three coupled weaknesses: 1)~sampling more from a single generator produces increasingly redundant candidates, 2)~existing pipelines apply one generic correction to every non-clean execution result, while runtime errors, timeouts, and empty results each indicate a different distance from correctness, and 3)~existing selectors rely on a single angle such as result-majority voting or pairwise SQL comparison, missing what other angles would have caught.
We present \system{}, which addresses all three weaknesses.
A difficulty-smoothing RL recipe trains SIRIUS-32B to generate diverse executable SQL candidates, paired with a generalist LLM that fills in gaps left by the specialist.
An execution-grounded lifecycle classifies each outcome and applies targeted repair before candidates re-enter the pool.
A confidence-gated hybrid selector combines execution-result agreement with pairwise SQL-form judgment, escalating only near-tied cases to a deterministic structural check.
\system{} reaches 75.88\% on \bird{} dev and 91.20\% on \spider{} test. Two of three generalist pairings surpass Agentar-Scale-SQL, the strongest published multi-candidate system on \bird{} dev.
\end{abstract}
\section{Introduction}
\label{sec:introduction}
Real-world databases expose persistent limits of LLM-based text-to-SQL: on \bird{}~\citep{li2023can}, the strongest leaderboard system, GPT-5.5-xhigh, reaches only 72.55\% execution accuracy on the test set.\footnote{Official BIRD leaderboard}
Schema heterogeneity, implicit constraints, and ambiguous value references generate errors that no single model reliably avoids~\citep{yu2018spider,lei2025spider,lin2020bridging,wang2025linkalign}.
To mitigate this, recent work converges on a common recipe: sample multiple SQL candidates per question and let voting over execution results filter individual errors~\citep{wangself,pourreza2025chase,liu2026xiyan,wang2025agentar}.

\begin{figure}[t]
    \centering
    \includegraphics[width=\columnwidth]{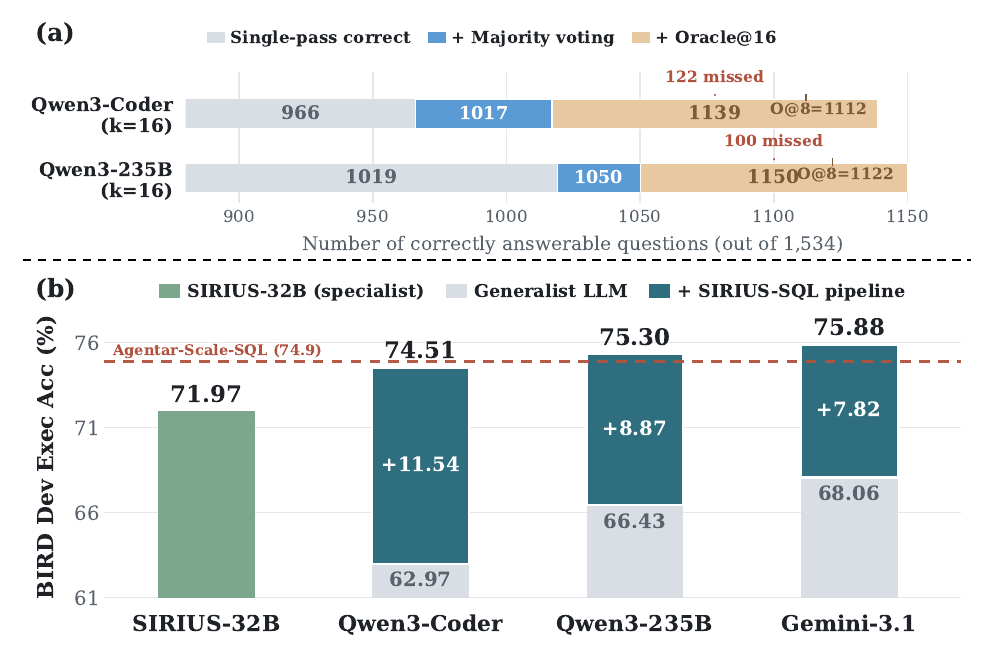}
    \vspace{-5ex}
    \caption{(a)~Coverage-to-accuracy gap on \bird{} dev.
    (b)~\system{} paired with three generalist LLMs, with the red dashed line marking Agentar-Scale-SQL.}
    \label{fig:intro}
     \vspace{-2ex}
\end{figure}

Figure~\ref{fig:intro}(a) decomposes each single-generator candidate set into what majority voting recovers and what remains reachable, exposing three structural limits.
These three limits trace to a common root: execution signals are systematically underused across training, repair, and selection.

\textbf{L1: Diminishing-return sampling.}
A single generator's candidate set hits diminishing returns as sampling grows~\citep{lee2025mcs,pourreza2025chase,liu2026xiyan}: on \bird{} dev, doubling the candidate budget from $k{=}8$ to $k{=}16$ adds only 27--28 reachable questions per generator\footnote{A question is reachable at $k$ (i.e., counted by Oracle@$k$, the bar length in Figure~\ref{fig:intro}(a)) if at least one of its $k$ sampled SQLs returns the correct execution result.}.
New samples largely repeat what the pool already contains, because each generator's outputs concentrate on the narrow region it was trained to prefer.
Existing multi-candidate systems address this at inference time by adding more prompts, decoding routes, or parallel experts~\citep{liu2026xiyan,pourreza2025chase,wang2025agentar,li2025deepeye}, leaving the training objectives that shape each individual source's output distribution largely untouched.

\textbf{L2: Wasted execution signal.}
The database itself returns rich diagnostic information whenever a candidate fails to produce a clean result~\citep{mao2024enhancing,xu2025ts,cen2025sqlfixagent,shinn2023reflexion}, yet existing pipelines apply one generic correction to every non-clean execution result.
A runtime error exposes a syntactic or type mismatch, a timeout points to an inefficient join or subquery, and an empty result indicates a semantic predicate drift, each pointing to a different distance from correctness.
Current pipelines either discard these candidates entirely or feed them to a single corrector with only a coarse label.

\textbf{L3: Voting--oracle gap.}
Even when the correct query is already in the candidate set, voting still misses it~\citep{pourreza2025chase,li2025deepeye}: 122 and 100 questions across the two single-generator pools in Figure~\ref{fig:intro}(a) have correct SQL in the pool yet are misclassified by majority voting.
The root cause is that selection requires joint judgment in two spaces: in the \emph{result space}, different SQL statements may return identical outputs despite different logic, and in the \emph{logic space}, the same correct intent admits many surface forms.
Existing selectors operate in only one space~\citep{pourreza2025chase,liu2026xiyan,li2025deepeye}, leaving the other as a blind spot.

We address these three limits with \system{}, which redesigns candidate construction, repair, and selection together.
\textbf{For L1}, \system{} trains SIRIUS-32B with a difficulty-smoothed two-stage RL recipe that broadens its candidate distribution while keeping each sample reward-validated, and pairs it with a generalist LLM for complementary coverage.
\textbf{For L2}, an execution-grounded lifecycle classifies each outcome into a type (runtime error, timeout, or empty result) and applies targeted repair before re-admitting the candidate.
\textbf{For L3}, a hybrid selector fuses tuple-level execution consensus with pairwise SQL quality assessment, and escalates only low-margin cases to a deterministic structural-support check. 
Our contributions are:
\begin{itemize}[leftmargin=*,itemsep=1pt,topsep=2pt]
    \item A \textbf{difficulty-smoothed two-stage RL recipe} that trains SIRIUS-32B to produce a broad yet reward-validated candidate distribution (\S\ref{sec:candidate-construction}).
    \item An \textbf{execution-grounded lifecycle} with type-specific repair for runtime errors, timeouts, and empty results (\S\ref{sec:lifecycle}).
    \item A \textbf{confidence-gated hybrid selector} that sums tuple-level consensus and pairwise SQL preference, with a deterministic AST gate as a zero-cost tie-breaker (\S\ref{sec:selection}).
\end{itemize}

\system{} reaches 75.88\% on \bird{} dev and 91.20\% on \spider{} test. Across three generalist partners (Figure~\ref{fig:intro}(b)), every pairing improves over its generalist baseline, and two surpass Agentar-Scale-SQL, the strongest published multi-candidate system on \bird{} dev.

\section{Related Work}

\noindent \textbf{Prompting and training-oriented generators.}
Early Text-to-SQL work studies schema-aware modeling, value grounding, intermediate representations, constrained decoding, and semantic evaluation~\citep{bogin2019representing,wang2020rat,lin2020bridging,zhong2020semantic,gan2021natural,qi2022rasat}.
Recent LLM-based systems improve a single generator through prompt design, task decomposition, task alignment, and self-correction~\citep{
gao2024text,pourreza2023din,qu2024before,xie2024decomposition,lee2025safe}.
Another line adapts SQL-oriented models through synthetic data, open-source specialization, self-taught reasoning, execution-driven bootstrapping, reward modeling, self-play fine-tuning, and execution-grounded reinforcement learning~\citep{yang2024synthesizing,li2024codes,
he2025star,zhang2025exesql,zhang2025spft,ma2025sqlr1,pourreza2025reasoningsql,yao2025arctic}.

\begin{figure*}[t]
    \centering
    \includegraphics[width=0.98\textwidth]{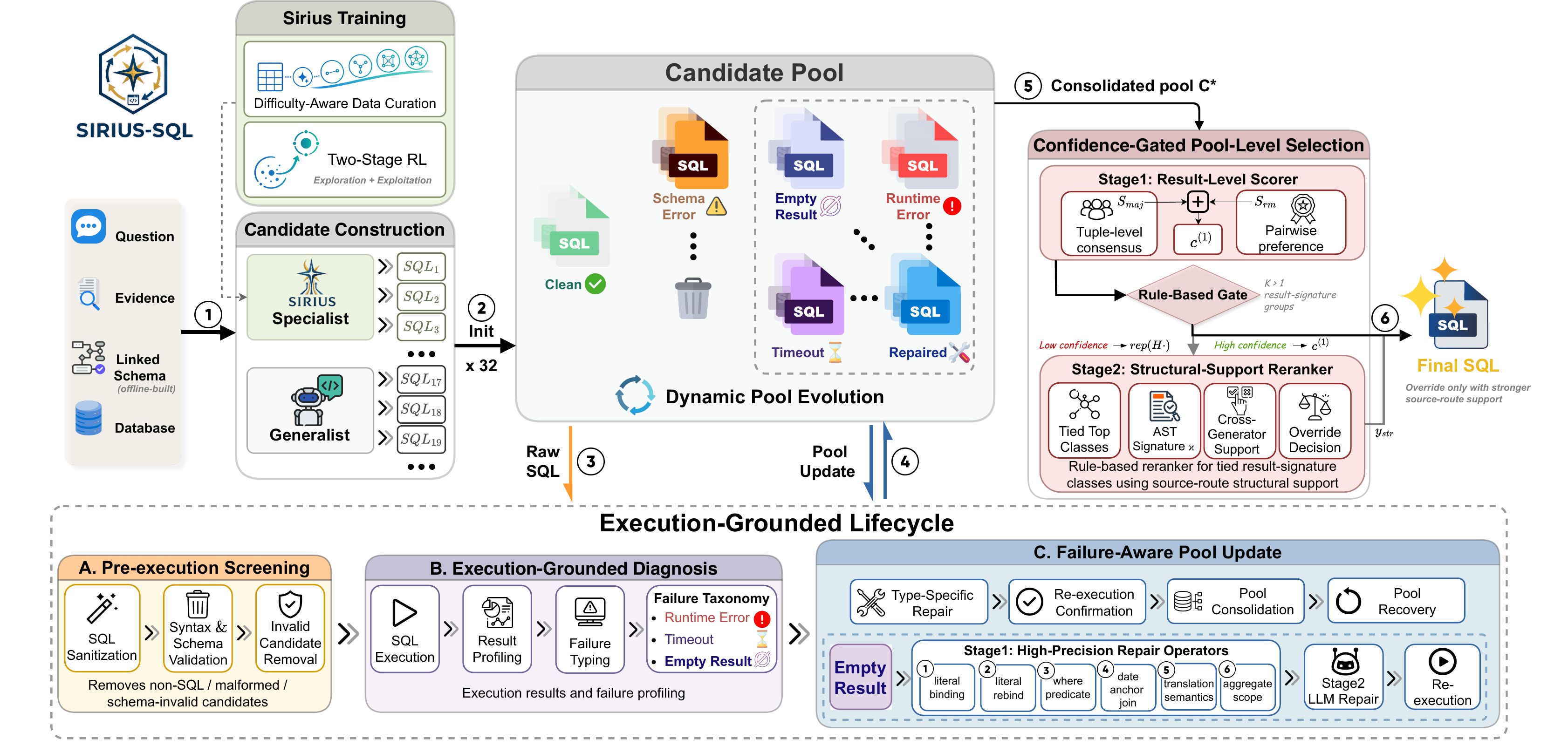}
    \vspace{-2ex}
    \caption{Overview of \system{}. A SQL specialist and a generalist LLM populate a shared candidate pool (\circled{1}\,\circled{2}; L1). An execution-grounded lifecycle screens, executes, and applies type-specific repair through Blocks~A--C (\circled{3}\,\circled{4}; L2). A confidence-gated selector scores the consolidated pool and escalates only low-margin ties to a deterministic structural-support reranker (\circled{5}\,\circled{6}; L3).}
    \label{fig:pipeline}
\end{figure*}

\noindent \textbf{Multi-candidate pipelines and selection.}
A second line of work moves beyond one-shot prediction by generating multiple candidates and organizing inference as a broader workflow.
MAC-SQL uses multi-agent collaboration~\citep{wang2025mac}, CHESS combines retrieval, schema pruning, generation, and verification~\citep{talaei2024chess}, Alpha-SQL formulates zero-shot generation as search~\citep{lialpha}, and Agentar-Scale-SQL emphasizes test-time orchestration~\citep{wang2025agentar}.
Closer to selection, MCS-SQL aggregates candidates from multiple prompts~\citep{lee2025mcs}, PET-SQL applies two-round refinement with cross-consistency~\citep{li2025pet}, CHASE-SQL and XiYan-SQL couple multi-path generation with pairwise candidate comparison as a final ranker after execution clustering~\citep{pourreza2025chase,liu2026xiyan}, and DeepEye-SQL introduces confidence-aware, execution-guided adjudication~\citep{li2025deepeye}.
Execution feedback has also been used for question rewriting, refinement, test-driven self-correction, and decomposed automatic repair~\citep{ mao2024enhancing}, while AST-level ranking has been explored for retrieval-augmented SQL generation~\citep{shen2024improving}.

\noindent \textbf{Positioning.}
Most multi-candidate Text-to-SQL systems expand the candidate set through prompting, search, or multiple generators, then apply a final voter, comparator, or verifier.
\system{} instead anchors every stage in execution feedback: an RLVR-trained specialist that produces broad yet executable candidates (L1), a typed-outcome lifecycle with re-execution before re-admission (L2), and a hybrid selector combining result-level, pairwise, and structural signals (L3).

\section{Method}
\subsection{Problem Formulation}
Given an input $x{=}(q,e,d)$ with question $q$, optional evidence $e$, and database $d$, \system{} predicts a SQL query $y$ from a shared candidate pool $\mathcal{C}(x){=}\{c_1,\ldots,c_n\}$ populated by complementary generators: a SQL specialist trained with verifiable execution rewards and an off-the-shelf generalist LLM, detailed in \S\ref{sec:candidate-construction}.
Let $\mathcal{E}(c,d)$ denote the execution outcome of candidate $c$ on $d$.
The pool is refined based on execution signals (\S\ref{sec:lifecycle}), and the final query is selected via a confidence-gated cascade over the refined pool (\S\ref{sec:selection}).

\subsection{System Overview}
Figure~\ref{fig:pipeline} summarizes \system{}.
Before generation, the system loads an offline schema prompt for every question and injects cached value hints. The schema prompt extends standard \texttt{CREATE TABLE} DDL with per-column comments and question-conditioned example values, so that ambiguous columns and unfamiliar value vocabularies are resolved before the generator sees them. Full construction details are in Appendix~\ref{sec:schema-details}.

\subsection{Source-Complementary Candidate Construction}
\label{sec:candidate-construction}
\label{sec:training}
Pools drawn from a single generator are bounded by that generator's training distribution: sampling more candidates from the same model does not reliably surface hypotheses outside what its distribution prefers (L1, \S\ref{sec:introduction})~\citep{lee2025mcs,pourreza2025chase}.
We therefore build the initial pool from two sources with complementary failure modes: a task-specialized generator $G_{\mathrm{sp}}$ and an external strong LLM $G_{\mathrm{ge}}$ serving as a generalist.

\noindent \textbf{Specialist.}
$G_{\mathrm{sp}}$ is the task-specialized source of SQL hypotheses, trained on a code-pretrained base with reinforcement learning from verifiable execution rewards (RLVR), a natural choice for SQL where each candidate's correctness is decidable by execution. Our training-side design choices are a \emph{difficulty-smoothed data mixture} and a \emph{two-stage DAPO schedule}.

\noindent \textit{Difficulty-smoothed data synthesis.}
Two design choices shape the training signal. On the reward side, we replace the binary executable-match reward with a three-valued scalar that additionally rewards executable-but-wrong rollouts, distinguishing partial progress from outright failure. On the data side, we form a difficulty-smoothed corpus by combining multiple SQL sources, filtering out noisy gold queries, and pruning prompts that the baseline already solves trivially, so that training compute concentrates on instances where the policy still has room to improve.
Reward values, filter stack, and pruning thresholds are in Appendix~\ref{sec:training-details}.

\noindent \textit{Two-stage DAPO with threshold-gated entropy.}
We adopt DAPO~\citep{yu2026dapo}, which builds on group-relative policy optimization for verifiable reasoning~\citep{shao2024deepseekmath}, as the policy-gradient backbone and augment it with a threshold-gated entropy regularizer that fires when policy entropy $H(\pi_\theta)$ falls below a target, counteracting entropy collapse without paying constant regularization:
\begin{equation}
\mathcal{L}_{\mathrm{total}}(\theta)
= \mathcal{L}_{\mathrm{DAPO}}(\theta)
+ w_{\mathrm{ent}}\,\mathcal{L}_{\mathrm{ent}}(\theta),
\label{eq:total-obj}
\end{equation}
\vspace{-1.2em}
\begin{equation}
w_{\mathrm{ent}}=
\begin{cases}
\beta, & H(\pi_\theta) < H_{\mathrm{target}},\\
0,     & \text{otherwise}.
\end{cases}
\label{eq:ent-gate}
\end{equation}
Training proceeds in two stages, with the objective shared. An \emph{exploration} stage keeps rollout space wide on hard prompts, and an \emph{exploitation} stage resumes from its checkpoint and anneals rollout width while preserving the dynamic-sampling constraint.
Hyperparameters, execution sandbox, and the DAPO surrogate are in Appendix~\ref{sec:training-details}.

\noindent \textbf{Generalist and pool union.}
$G_{\mathrm{ge}}$ is an off-the-shelf general-purpose LLM that provides semantic coverage for long-tail phrasing and implicit constraints, prompted with 5-shot in-context demonstrations while $G_{\mathrm{sp}}$ uses its task-tuned prompt format.
$G_{\mathrm{sp}}$ contributes high-consensus clusters of execution-equivalent rewrites, while $G_{\mathrm{ge}}$ contributes broader coverage of cases that $G_{\mathrm{sp}}$ may miss.
At inference time, both generators are sampled under a fixed budget, using multiple prompt routes within each source, and the resulting hypotheses are merged into a single pool.
The construction objective is reduced error correlation rather than diversity alone: candidates from different sources should expose different failure modes, and downstream stages retain source and route metadata as provenance signals.

\subsection{Execution-Grounded Candidate Lifecycle}
\label{sec:lifecycle}
Execution is used not only as a final verifier but also as the organizing signal for pool management.
The lifecycle is organized into three blocks (Figure~\ref{fig:pipeline}): (A)~Pre-execution Screening, (B)~Execution-Grounded Diagnosis, and (C)~Failure-Aware Pool Update.
The key design choice is to front-load deterministic screening, spend repair budget on execution-near hypotheses, and recycle unresolved failures as negative feedback only when the pool collapses. 
Appendix~\ref{sec:lifecycle-details} details the high-level procedure, repair path, the empty-result operator catalog, and the post-repair rules.

\noindent \textbf{Pre-execution screening.}
We first apply rule-based cleanup to common LLM artifacts and then filter the pool with lightweight syntax checks and
\texttt{EXPLAIN}-backed schema validation.
Candidates that fail this stage are removed from the pool before execution. This front-loaded filter is deliberate: obviously malformed SQL (such as comment-prefixed output, missing \texttt{FROM}/\texttt{SELECT} clauses, bracket or quote mismatches, or references to nonexistent tables and columns) should not consume downstream repair budget. Sanitization rules and validator details are fully specified in Appendix~\ref{sec:lifecycle-block-a}.

\noindent \textbf{Execution-grounded diagnosis.}
Surviving candidates are executed on the database and classified by a four-way taxonomy (\texttt{clean}, \texttt{runtime}, \texttt{timeout}, \texttt{empty}), where \texttt{empty} denotes valid queries returning zero rows, \texttt{runtime} indicates an engine exception, and \texttt{timeout} is caught by a per-query wall-clock budget.
The taxonomy is load-bearing because different classes require structurally different repairs.

\noindent \textbf{Failure-aware repair and pool consolidation.}
Every accepted rewrite is \emph{immediately re-executed} before it can re-enter the pool, and each path is shaped by what the database outcome reveals. A runtime error pinpoints a concrete schema or syntax issue, and the engine's exception text is fed verbatim to the corrector. A timeout indicates a semantically plausible candidate with an inefficient query plan, so the rewrite preserves intent and restructures join order or subquery shape. An empty result indicates that the SQL ran but missed the answer (often value-binding errors or over-restrictive predicates), and six high-precision structural operators try local recovery before a diagnosis-conditioned LLM fallback.
After repair, a deterministic projection-normalization pass rewrites noisy \texttt{SELECT} lists into cleaner forms, for example, dropping unrequested numbered variants such as \texttt{Email2}/\texttt{Email3} or canonicalizing known column bundles. Full operator definitions and post-repair rules are in Appendix~\ref{sec:lifecycle-block-c}--\ref{sec:lifecycle-postrepair}.

\noindent \textbf{Pool recovery.}
If post-repair filtering exhausts the pool, a recovery generator receives condensed negative feedback from unresolved candidates, rule-filtered SQLs, and matched failure tags, then regenerates a replacement pool (Appendix~\ref{sec:lifecycle-postrepair}).

\subsection{Confidence-Gated Hybrid Selection}
\label{sec:selection}
Given the consolidated pool $\mathcal{C}^{(*)}$ (Figure~\ref{fig:pipeline}, right), the selector operates in two stages that correspond to the two boxes in the selection panel.
\textbf{Stage~1} (Result-Level Scorer) computes a single coarse score $S_{\mathrm{coarse}}{=}S_{\mathrm{maj}}{+}S_{\mathrm{rm}}$ per candidate and takes its argmax as the default winner $c^{(1)}$.
A rule-based confidence gate then decides whether $c^{(1)}$ is sufficiently reliable. If not, \textbf{Stage~2} (Structural-Support Reranker) checks whether another tied class has strictly stronger cross-source structural support and overrides the default winner only in that case.

\noindent \textbf{Stage~1: Result-level scorer.}
Let $\mathcal{C}^{+}{=}\{c\in\mathcal{C}^{(*)}\mid\delta(c){=} \texttt{clean}\}$ be the successfully executed candidates.
Stage~1 combines two complementary result-level signals over $\mathcal{C}^{+}$ and picks the argmax of their sum.

\noindent \textit{Tuple-level consensus $S_{\mathrm{maj}}$.}
We decompose each candidate's result set into cell-level triples $\mathcal{T}(c){=}\{(i,\mathit{col}_j,v_{ij})\}$, count the cross-pool
frequency of each triple $f(t){=}|\{c'\!\in\!\mathcal{C}^{+}\mid t\in\mathcal{T}(c')\}|$, and average these frequencies with \texttt{null}-valued cells down-weighted:
\begin{equation}
S_{\mathrm{maj}}(c)
=\frac{1}{|\mathcal{T}(c)|}
 \!\sum_{t\in\mathcal{T}(c)}\!
 f(t)\,\mathbf{1}[v(t){\neq}\texttt{null}].
\label{eq:smaj}
\end{equation}
Empty-result candidates receive $S_{\mathrm{maj}}{=}0$ and are thus excluded from the consensus (derivation in Appendix~\ref{sec:selection-consensus}).

\noindent \textit{Position-debiased pairwise preference $S_{\mathrm{rm}}$.}
We partition $\mathcal{C}^{+}$ into equivalence classes by result signature $\sigma(c){=}\mathrm{frozenset}(\text{rows of }c)$.\footnote{Result-signature grouping lets syntactically different SQL queries share evidence when they return the same tuple set.}
We pick one representative per class and query an LLM judge on every pair of representatives in \emph{both orderings} to cancel position bias~\citep{zheng2023judging,wang2024large}.
Each representative accumulates a wins-minus-losses score broadcast to class members (Appendix~\ref{sec:selection-elo}).

Stage~1 then returns
\begin{equation}
c^{(1)}=\arg\max_{c\in\mathcal{C}^{+}}
\bigl[S_{\mathrm{maj}}(c)+S_{\mathrm{rm}}(c)\bigr],
\label{eq:coarse}
\end{equation}
with single-signal variants (\emph{freq}, \emph{majority voting}, \emph{majority-refine}, \emph{RM}) used in the RQ5 ablation defined in Appendix~\ref{sec:selection-variants}.
Besides the default merged score, the implementation also materializes an auxiliary consensus-refine statistic used only by the confidence gate. Appendix~\ref{sec:selection-details} gives the implementation-level definition.

\noindent \textbf{Rule-based confidence gate.}
The gate is a strict ambiguity detector over Stage-1.
Let $K{=}|[\mathcal{C}^{+}]_\sigma|$ be the number of result-signature classes.
The gate fires if and only if the top-2 gaps under $S_{\mathrm{coarse}}$ (merge), an auxiliary consensus-refine score $S_{\mathrm{ref}}$ (refine), and $S_{\mathrm{rm}}$ (RM) are \emph{all} zero, while $1{<}K{\le}K_{\max}$ and the largest class has size ${\le}T_{\max}$ (values in Appendix~\ref{sec:selection-details}).
The zero-margin conjunction ensures escalation only when every Stage-1 signal is simultaneously tied, and the cardinality bounds exclude pools that are too fragmented or already dominated, both of which are well-handled by $c^{(1)}$.
Empirically, this restricts Stage~2 to a small fraction of ambiguous questions.

\noindent \textbf{Stage~2: Structural-support reranker.}
When the gate fires, a deterministic, LLM-free reranker is applied to the result-signature classes tied at the maximum Stage-1 coarse score.
Let $\mathcal{G}_{\mathrm{top}}$ denote this tied set, and let $\hat{c}_g$ be the representative candidate of class $g\in\mathcal{G}_{\mathrm{top}}$.
Each candidate is parsed into an AST-level structural signature $\kappa(c)$ capturing its logical skeleton: table set, projection patterns, projected/predicate columns, aggregation/order markers, and a coarse limit bucket (full feature list in Appendix~\ref{sec:selection-signature})~\citep{shen2024improving}.
The reranker then measures \emph{cross-source support}: how many distinct provenance groups $\mathrm{srcgrp}(c)$ (generator--route pairs, with model-level fallback) contribute a candidate sharing the same structural signature, so that convergence from multiple independent sources outweighs support concentrated within one generator.
For each tied class, let
\begin{align}
\mathcal{S}_g
&= \{c\!\in\!\mathcal{C}^{(*)}\mid\kappa(c){=}\kappa(\hat{c}_g)\},
\notag\\
\mathrm{supp}(g)
&= \bigl|\{\mathrm{srcgrp}(c)\mid c\in\mathcal{S}_g\}\bigr|,
\notag\\
g^*
&=\arg\max_{g\,\in\,\mathcal{G}_{\mathrm{top}}}
\mathrm{supp}(g),
\label{eq:structural}
\end{align}
where $\mathcal{S}_g$ is the support set of candidates sharing the same structural signature as representative $\hat{c}_g$.

\noindent \textbf{Final decision.}
Let $g^{(1)}$ be the result-signature class containing $c^{(1)}$.
The structural branch overrides the Stage-1 winner only when $\mathsf{gate}(x)$ holds and a competing tied class receives strictly stronger support, i.e.\ $\mathrm{supp}(g^*)>\mathrm{supp}(g^{(1)})$. Otherwise $y{=}c^{(1)}$.
The only LLM calls used in selection are those already required to compute $S_{\mathrm{rm}}$ in Stage~1.
Algorithm~\ref{alg:selection} in Appendix~\ref{sec:selection-details} gives full procedure.

\section{Experimental Setup}
\label{sec:experimental-setup}
We evaluate on \bird{}~\citep{li2023can} (1,534 dev questions, 95 databases) as the primary benchmark and \spider{}~\citep{yu2018spider} (2,147 test questions) for generalization. All ablations run on \bird{} dev. We report execution accuracy (EX) with a difficulty breakdown on \bird. Evaluation-protocol details, including benchmark-specific execution matching rules, are in Appendix~\ref{sec:appendix-evaluation}.

All main runs use the default enriched schema context. Schema-variant ablations are reported in Appendix~\ref{sec:schema-variant-results}. The specialist is SIRIUS-32B (Appendix~\ref{sec:training-details}). Generalist partners are Qwen3-235B-Instruct-2507~\citep{yang2025qwen3}, Gemini-3.1-Pro-Preview, and Qwen3-Coder-30B-A3B-Instruct. Each question receives up to 32 candidates across generators. Serving, timeout, decoding, and backbone-matched baseline details are in Appendices~\ref{sec:appendix-serving} and \ref{sec:appendix-baselines}.

\section{Results and Analysis}
\label{sec:results}

We organize our study around five RQs.
RQ1 (\S\ref{sec:rq1-main-results}) compares \system{} against published systems. RQ2--RQ5 are ablations isolating each design decision on \bird{} dev: difficulty-smoothed data synthesis (\S\ref{sec:ablation-training}), complementary candidate construction (\S\ref{sec:ablation-candidate-construction}), failure-aware repair (\S\ref{sec:ablation-lifecycle}), and confidence-gated selection (\S\ref{sec:ablation-selection}).

\subsection{RQ1: How Does \system{} Compare to Prior Systems?}
\label{sec:rq1-main-results}

\begin{table*}[t]
    \centering
    \caption{Main results. Execution accuracy on \bird{} dev (1,534 questions, with difficulty breakdown) and \spider{} 1.0 test (2,147 questions, gold execution).
    $^{\dagger}$: source paper.
    $^{\S}$: public leaderboard.
    Unmarked: our runs.}
    \vspace{-1ex}
    \label{tab:main-results}
    \scriptsize
    \setlength{\tabcolsep}{12pt}
    \renewcommand{\arraystretch}{1}
    \begin{tabular}{@{}llccccc@{}}
        \toprule
        & & \multicolumn{4}{c}{\bird} & \spider \\
        \cmidrule(lr){3-6}\cmidrule(lr){7-7}
        Method & Model & Simple $\uparrow$ & Moderate $\uparrow$ & Challenging $\uparrow$ & Exec. $\uparrow$ & Exec. $\uparrow$ \\
        \midrule
        GPT-4o$^{\dagger}$ & GPT-4o & -- & -- & -- & 61.9 & 83.2 \\
        Qwen3-Coder & Qwen3-Coder-30B & 69.73 & 54.31 & 47.59 & 62.97 & 85.10 \\
        Qwen3-235B & Qwen3-235B & 72.54 & 58.41 & 53.10 & 66.43 & 86.21 \\
        Gemini-3.1 Pro & Gemini-3.1 Pro & 74.27 & 60.78 & 51.72 & 68.06 & 88.02 \\
        \midrule
        DIN-SQL$^{\S}$ & GPT-4 & -- & -- & -- & 50.72 & 85.3 \\
        DAIL-SQL$^{\S}$ & GPT-4 & -- & -- & -- & 54.76 & 86.6 \\
        MCS-SQL$^{\dagger}$ & GPT-4 8K & -- & -- & -- & 63.4 & 89.6 \\
        OmniSQL$^{\dagger}$ & OmniSQL-32B & -- & -- & -- & 67.0 & 89.8 \\
        OpenSearch-SQL$^{\dagger}$ & GPT-4o & -- & -- & -- & 69.3 & 87.1 \\
        Arctic-Text2SQL-R1$^{\dagger}$ & Arctic-Text2SQL-R1-32B & -- & -- & -- & 70.5 & 88.7 \\
        DeepEye-SQL & Qwen3-235B & 76.22 & 64.87 & 59.31 & 71.12 & 87.61 \\
        Reasoning-SQL$^{\dagger}$ & Qwen2.5-Coder-14B & -- & -- & -- & 72.29 & 81.43 \\
        CHASE-SQL$^{\dagger}$ & Gemini 1.5 & -- & -- & -- & 73.01 & 87.6 \\
        XiYan-SQL$^{\dagger}$ & XiYanSQL-32B + GPT-4o & -- & -- & -- & 73.34 & 89.65 \\
        DeepEye-SQL$^{\dagger}$ & Qwen3-Coder-30B & -- & -- & -- & 73.53 & 89.8 \\
        LongData-SQL$^{\S}$ & -- & -- & -- & -- & 74.32 & -- \\
        Agentar-Scale-SQL$^{\S}$ & Agentar-32B + Gemini & -- & -- & -- & 74.90 & -- \\
        \midrule
        SIRIUS & SIRIUS-32B & 76.22 & 67.67 & 58.62 & 71.97 & 89.99 \\
        SIRIUS-SQL & SIRIUS-32B + Qwen3-Coder-30B & 79.24 & 68.97 & 62.07 & 74.51 & 89.85 \\
        SIRIUS-SQL & SIRIUS-32B + Qwen3-235B & 80.00 & 69.40 & 64.14 & 75.30 & 90.41 \\
        SIRIUS-SQL & SIRIUS-32B + Gemini-3.1 Pro & \textbf{80.54} & \textbf{70.26} & \textbf{64.14} & \textbf{75.88} & \textbf{91.20} \\
        \bottomrule
    \end{tabular}
    \vspace{-1ex}
\end{table*}

\system{} outperforms all baselines on both benchmarks and generalizes across generalist partners (Table~\ref{tab:main-results}).
On \bird{} dev, SIRIUS-32B + Gemini-3.1 Pro reaches $75.88$ EX, $+0.98$ above Agentar-Scale-SQL ($74.90$).
Over the best single-pass generalist (Gemini-3.1 Pro), \system{} improves Moderate by $+9.48$ and Challenging by $+12.42$.
All three pairings improve over both their specialist alone ($\geq{+}2.54$ EX) and their corresponding generalist baseline ($\geq{+}7.82$ EX), and their ordering tracks each generalist's standalone strength, supporting partner-robust pipeline gains.
On \spider{} test, the same configuration transfers without retuning to $91.20$ EX, above all baselines.

\subsection{RQ2: Does Difficulty-Smoothed Data Synthesis Help Specialist RL Training?}
\label{sec:ablation-training}

We isolate the effect of difficulty-smoothed data synthesis (\S\ref{sec:candidate-construction}) at 7B scale: only the data synthesis procedure is toggled, with all other training settings held fixed (Appendix~\ref{sec:training-details}).

\begin{figure}[t]
    \centering
    \includegraphics[width=\columnwidth]{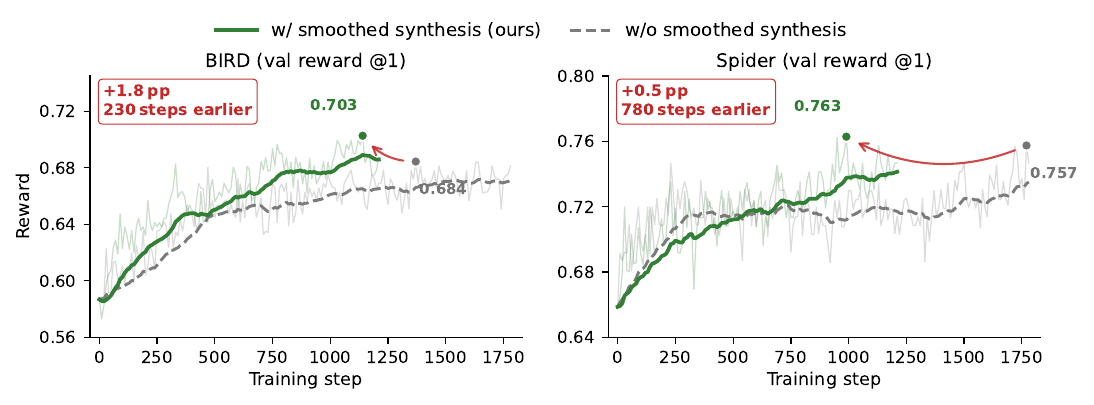}
    \vspace{-5ex}
    \caption{Training-source validation reward@1 during 7B specialist RL, with and without difficulty-smoothed data synthesis.}
    \label{fig:training-curves}
\end{figure}

Figure~\ref{fig:training-curves} shows that difficulty-smoothed synthesis reaches a higher peak validation reward \emph{and} reaches it earlier on both benchmarks: peak $0.703$ on \bird{} ($+1.8$\,pp, $230$ steps earlier) and $0.763$ on \spider{} ($+0.5$\,pp, $780$ steps earlier).
Smoothing concentrates the training corpus on prompts where the baseline is genuinely uncertain, so each RL batch carries a more informative reward signal throughout training.
We then apply the same recipe to train SIRIUS-32B (specialist for all main runs); the 32B curve and additional diagnostics are in Appendix~\ref{sec:training-details} and \ref{sec:training-critic}.

\subsection{RQ3: Does Complementary Candidate Construction Help?}
\label{sec:ablation-candidate-construction}

RQ3 asks whether candidate construction makes correct SQL both \emph{reachable} (measured by Oracle@$k$) and \emph{selectable} (proxied by majority voting), rather than merely increasing the sample count. We compare variants on \bird{} dev under a matched candidate budget: each hybrid variant produces $N_s{=}16$ specialist samples (task-specific prompt) and $N_g{=}16$ generalist samples (5-shot ICL prompt), and single-source variants sample one generator $N_s{+}N_g$ times. All variants share the same lifecycle and selector.

\begin{table}[t]
    \centering
    \caption{Candidate construction ablation (RQ3). \emph{5-shots}: $k{=}1$ greedy decoding with 5 in-context demonstrations. \emph{matched}: $k{=}N_s{+}N_g$ samples drawn from the single source under the same total budget as Hybrid.}
    \label{tab:candidate-construction}
    \vspace{-1ex}
    \footnotesize
    \setlength{\tabcolsep}{3pt}
    \resizebox{\columnwidth}{!}{%
    \begin{tabular}{llcccc}
        \toprule
        \# & Variant & $k$ & Oracle@$k$ $\uparrow$ & Maj Voting $\uparrow$ & Final Exec.\ $\uparrow$ \\
        \midrule
        V1 & Qwen3-coder (5-shots)   & 1 & - & - & 66.56\drop{7.95} \\
        V2 & Qwen3-235B (5-shots)   & 1 & - & - & 69.30\drop{5.21} \\
        V3 & XiYanSQL-32B   & 1 & - & - & 64.28\drop{10.23} \\
        V4 & Agentar-32B   & 1 & - & - & 70.66\drop{3.85} \\
        V5 & SIRIUS-32B   & 1 & - & - & 71.97\drop{2.54} \\
        \midrule
        V6 & Qwen3-Coder only  & $N_s{+}N_g$ & 80.18 & 71.38 & 72.08\drop{2.43} \\
        V7 & SIRIUS-32B only   & $N_s{+}N_g$ & 78.42 & 73.14 & 73.27\drop{1.24} \\
        \midrule
        V8 & Qwen3-235B+Qwen3-Coder  & $N_s{+}N_g$ & 80.52 & 72.26 & 72.41\drop{2.10} \\
        V9 & XiYanSQL-32B+Qwen3-Coder  & $N_s{+}N_g$ & 83.25 & 72.69 & 72.88\drop{1.63} \\
        V10 & Agentar-32B+Qwen3-Coder  & $N_s{+}N_g$ & 82.59 & 74.05 & 73.92\drop{0.59} \\
        \midrule
        V11 & \textbf{Our (SIRIUS-32B+Qwen3-Coder)}      & $N_s{+}N_g$ & \textbf{82.66} & \textbf{74.12} & \textbf{74.51} \\
        \bottomrule
    \end{tabular}%
    }
\end{table}

Scaling Qwen3-Coder alone (V6) reaches a higher Oracle than scaling SIRIUS-32B alone (V7) ($80.18$ vs.\ $78.42$), but its majority vote is lower ($71.38$ vs.\ $73.14$).
The two strengths complement each other: the generalist often reaches a correct SQL without enough within-pool support, while SIRIUS-32B has lower reachability but stronger consensus among its sampled candidates.
Holding the generalist fixed, replacing the second generalist (V8: Qwen3-235B+Coder, $72.41$) with our RL-trained specialist (V11) adds $+2.10$ EX and lifts Oracle by $+2.14$pp, isolating the value of having an SQL-specialized, RL-trained source in the pool.
\begin{figure}[t]
    \centering
    \includegraphics[width=\columnwidth]{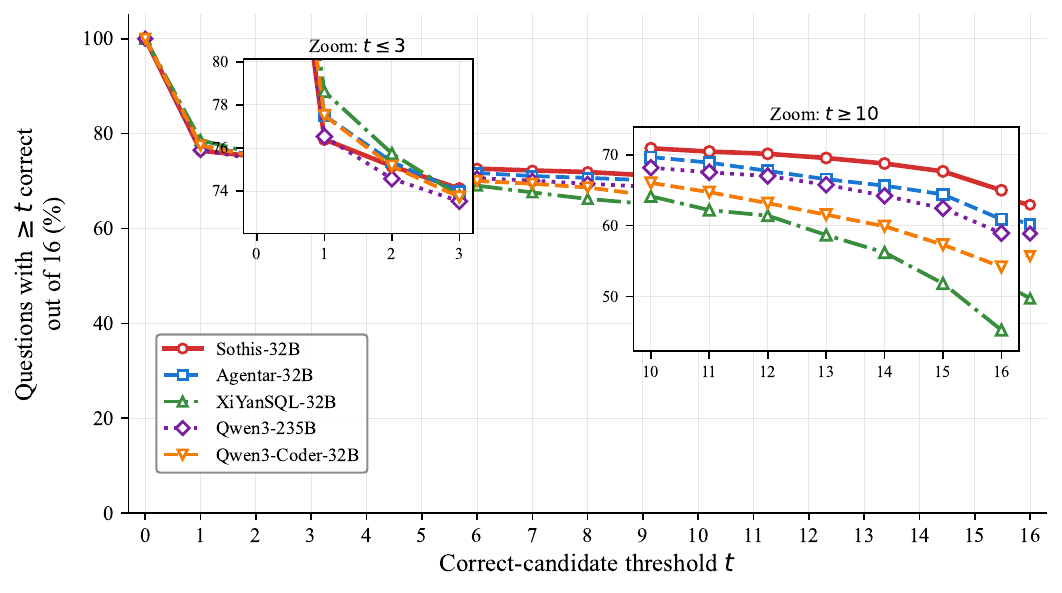}
    \vspace{-2em}
    \caption{Correct-candidate concentration in single-source $16$-sample pools.
    Each curve plots the fraction of questions with at least $t$ execution-correct candidates.
    The point $t{=}1$ is Oracle@16, while a larger $t$ indicates stronger support for voting and selection.}
    \label{fig:correctness-cdf}
\end{figure}

V8--V10 replace SIRIUS-32B with concurrent specialists. Their Oracle ceilings are comparable ($80.52$--$83.25$ vs.\ $82.66$), but SIRIUS-32B gives the best majority voting and final EX, and Figure~\ref{fig:correctness-cdf} shows why.
XiYanSQL-32B starts higher at $t{=}1$ ($78.62\%$ vs.\ $76.40\%$) but collapses to $45.24\%$ at $t{=}16$, while SIRIUS-32B stays at $70.93\%$ at $t{\ge}10$ and $64.99\%$ at $t{=}16$.
Thus, SIRIUS-32B supplies fewer isolated hits and more high-support correct clusters, which follows from the RLVR training in \S\ref{sec:candidate-construction}: rewards depend only on execution correctness, so the trained specialist tends to sample multiple SQL rewrites that share the same correct execution result, anchoring majority voting once merged with a broad generalist.
\subsection{RQ4: Does Failure-Aware Repair Beat Unified Correction?}
\label{sec:ablation-lifecycle}

The full type-specific lifecycle outperforms every ablated variant by $0.13$--$0.71$pp (Table~\ref{tab:lifecycle-ablation}). We fix the default \emph{Hybrid pool} (SIRIUS-32B + Qwen3-Coder unless noted otherwise) with exact-result majority voting as the selector, isolating the lifecycle alone.

\begin{table}[t]
    \centering
    \caption{Lifecycle ablation (RQ4) with selector held at exact-result majority voting.}
    \label{tab:lifecycle-ablation}
    \vspace{-1ex}
    \footnotesize
    \setlength{\tabcolsep}{3pt}
    \resizebox{\columnwidth}{!}{%
    \begin{tabular}{lcc}
        \toprule
        Variant & Exec.\ $\uparrow$ & Oracle@$k$ $\uparrow$ \\
        \midrule
        w/o lifecycle                          & 74.32\drop{0.71} & 81.49 \\
        w/o pre-execution screening            & 74.90\drop{0.13} & 82.40 \\
        w/o failure typing (drop failures)     & 74.77\drop{0.26} & 81.55 \\
        w/o failure typing (coarse LLM repair) & 74.71\drop{0.32} & 81.81 \\
        w/o post-repair filtering              & 74.77\drop{0.26} & 82.14 \\
        \midrule
        \textbf{\system{} (full lifecycle)}    & \textbf{75.03} & \textbf{82.27} \\
        \bottomrule
    \end{tabular}%
    }
\end{table}

The most informative row is \emph{coarse LLM repair} ($-0.32$), which routes non-clean candidates through a generic corrector that sees only the failed SQL. It underperforms even dropping them outright ($-0.26$): without the typed signal, the corrector treats a timeout from a costly join the same as a silently empty predicate, and produces plausibly-clean but still-wrong rewrites that outvote the correct ones. The typed dispatch in \S\ref{sec:lifecycle} routes each outcome to a structurally different repair path, which is what the coarse-LLM ablation removes.

\begin{figure}[t]
    \centering
    \vspace{-1ex}
    \includegraphics[width=\columnwidth]{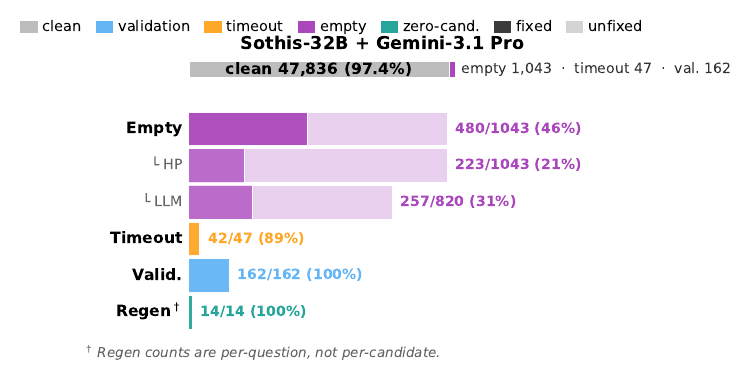}
    \vspace{-4ex}
    \caption{Per-mechanism repair rates on the SIRIUS-32B~+~Gemini-3.1~Pro pool ($49{,}088$ candidates on \bird{} dev). Per-model breakdowns in Appendix Fig.~\ref{fig:failure-breakdown-appendix}.}
    \label{fig:failure-breakdown-main}
    \vspace{-1ex}
\end{figure}

Fig.~\ref{fig:failure-breakdown-main} attributes the $+0.71$pp gain to where each lifecycle component is actually doing work. The three components consume non-overlapping slices of the candidate population. Pre-execution screening is responsible for the near-zero runtime-error bar: schema-mismatch candidates are filtered before reaching execution-grounded diagnosis, so almost no runtime errors survive to the typed-repair stage. The cross-check is the second row of Table~\ref{tab:lifecycle-ablation}: removing screening exposes $92$ runtime-error candidates to the back-end repair path, resulting in a $-0.13$pp drop in Final Exec.\ The cost-oriented prompt for timeouts (\S\ref{sec:lifecycle}) resolves nearly all timeout candidates, since such candidates typically differ from correct ones only in join efficiency. Two-stage empty-result repair is the hardest category: the high-precision (HP) structural operators (Stage~1) and the LLM fallback (Stage~2) together recover close to half of all empty-result candidates, with the remainder dominated by value-binding and projection mismatches.

\subsection{RQ5: Does Confidence-Gated Selection Beat Single-Signal Selectors?}
\label{sec:ablation-selection}

Two complementary signals already account for nearly all the gap between single-signal selectors and the full system, while a deterministic gate provides a zero-cost robustness check on the residual ties (Table~\ref{tab:selection-ablation}).
We fix the Hybrid pool with Gemini-3.1 Pro as a generalist (the configuration that achieves the headline $75.88$ EX) and the full failure-aware lifecycle, varying only the selector. Selector variants and RM training are detailed in Appendix~\ref{sec:selection-variants} and \ref{sec:rm-training}.

\begin{table}[t]
    \centering
    \caption{Selector ablation (RQ5). Variant definitions in Appendix~\ref{sec:selection-variants}. \emph{merge} = $S_{\mathrm{maj}} + S_{\mathrm{rm}}$ (\S\ref{sec:selection}).}
    \label{tab:selection-ablation}
    \vspace{-1ex}
    \footnotesize
    \setlength{\tabcolsep}{3pt}
    \resizebox{\columnwidth}{!}{%
    \begin{tabular}{lcccc}
        \toprule
        Selector & Simple $\uparrow$ & Moderate $\uparrow$ & Challenging $\uparrow$ & Exec.\ $\uparrow$ \\
        \midrule
        freq                                    & 79.24 & 70.04 & 63.45 & 74.97\drop{0.91} \\
        majority voting                         & 79.57 & 70.04 & 62.07 & 75.03\drop{0.85} \\
        majority-refine                         & 79.35 & 69.61 & 63.45 & 74.90\drop{0.98} \\
        RM                                      & 79.24 & 69.40 & 58.62 & 74.32\drop{1.56} \\
        merge                                   & 80.32 & 70.04 & 64.14 & 75.68\drop{0.20} \\
        \midrule
        \textbf{\system{} (merge~+~gate)}       & \textbf{80.54} & \textbf{70.26} & \textbf{64.14} & \textbf{75.88} \\
        \bottomrule
    \end{tabular}%
    }
    \vspace{-1ex}
\end{table}

Most of the gain comes from \emph{merge}, the simple sum $S_{\mathrm{maj}}{+}S_{\mathrm{rm}}$. The three voting variants (freq, majority, majority-refine) cluster within $0.13$pp of each other at $74.9$--$75.0$, indicating that result-level consensus on its own has saturated. RM alone is the weakest single signal at $74.32$, because pairwise preference ranks representatives within a cluster of execution-equivalent candidates rather than discriminating across clusters. Merge combines the two without learned weighting and lifts EX to $75.68$ ($+0.65$ over majority voting).

The gate fires only when the top-two gaps in all three scores (merge, consensus-refine, RM) vanish simultaneously \emph{and} multiple result-signature classes are present within size limits (\S\ref{sec:selection}). The gate then reverts to AST-level structural support without additional LLM calls. The resulting $+0.20$ EX affects only a small minority of dev questions and is Pareto-improving: same compute, higher accuracy, no parameters tuned across pools.

\section{Conclusion}
We address three structural limitations of multi-candidate Text-to-SQL voting (L1--L3) by anchoring every stage of the pipeline in execution feedback. \system{} pairs an RLVR-trained SQL specialist with a generalist LLM, refines the resulting pool through an execution-grounded lifecycle with type-specific repair, and selects the final SQL with a hybrid selector that resolves most questions by pooled result-level evidence and escalates only low-margin cases to a deterministic structural check.
\system{} reaches $75.88\%$ on \bird{} dev and $91.20\%$ on \spider{} test. Two of three generalist pairings surpass Agentar-Scale-SQL, and all three improve over their generalist baselines.


\section{Limitations}
\system{} requires both a domain-specialized RL-trained model and a strong generalist LLM. Deployments that lack either will not benefit equally.
The empirical residual headroom of the lifecycle focuses on \emph{empty-result} candidates whose errors are dominated by value-binding and projection mismatches (Fig.~\ref{fig:failure-breakdown-main}), where our type-specific operators recover roughly half of these candidates, but the remainder remains a hard frontier.
Finally, comparisons to published systems rely on their reported numbers and may differ in the retrieval pipeline, schema linking, or backbone. We therefore use the backbone-matched controls in RQ3 as the primary evidence for the design choices.

\section*{Ethics Statement}
This work focuses on benchmark evaluation of Text-to-SQL systems and does not involve human subjects, private data, or dual-use concerns.
All experiments use publicly available benchmarks and models.
We use publicly available benchmarks and models according to their intended research use and applicable licenses or terms of use.
Any released model checkpoints and derived artifacts will be distributed under licenses compatible with the applicable licenses and terms of the underlying resources.
\bibliography{custom}

\appendix
\newpage

\clearpage

\section{Appendix Overview}
This appendix is organized to mirror the evidence chain in the main paper.
Appendix~\ref{sec:experiment-details-appendix} first fixes the experimental
protocol, baseline settings, and input schema context used across all runs.
Appendix~\ref{sec:training-details} then expands L1, covering specialist
training, difficulty smoothing, the DAPO objective, the execution sandbox, and
training diagnostics.
Appendix~\ref{sec:lifecycle-details} expands L2, giving the execution-grounded
candidate lifecycle, its diagnostic taxonomy, repair operators, and additional
failure-breakdown evidence.
Appendix~\ref{sec:selection-details} expands L3, including tuple-level
consensus, score-family ablations, position-debiased pairwise preference, the
pairwise reward model, and structural-signature features.
Algorithms~\ref{alg:lifecycle} and~\ref{alg:selection} are placed inside their corresponding L2 and L3 sections. Appendix~\ref{sec:appendix-discussion} collects extended discussion.
The appendix figures are marked at the point where they support the main
experimental claims: Fig.~\ref{fig:difficulty-distribution} and
Fig.~\ref{fig:training-curves-32b} support L1 training, Fig.~\ref{fig:training-curves-critic}
adds the critic-reward diagnostic for RQ2, and
Fig.~\ref{fig:failure-breakdown-appendix} expands the L2 failure analysis.

\section{Experimental Protocol and Input Context}
\label{sec:experiment-details-appendix}

This section collects the implementation, evaluation, baseline, and input-context details that are summarized in compact form in \S\ref{sec:experimental-setup} of the main paper.

\subsection{Serving, Timeouts, and Decoding Profiles}
\label{sec:appendix-serving}

Open-weight models (Sothis-32B, Qwen3-235B-Instruct-2507, Qwen3-Coder-30B-A3B-Instruct) are served through OpenAI-compatible endpoints with vLLM 0.11~\citep{kwon2023efficient}.
Gemini-3.1-Pro-Preview is accessed through its hosted API.
Query execution uses two timeout budgets: $120$ seconds during the candidate lifecycle, and $30$ seconds during selection-time re-execution.
The tighter selection-time budget rejects candidates whose earlier acceptance depended on a transient long run, while the lifecycle budget is loose enough to avoid penalizing expensive but correct queries during repair.
Each question receives $16$ candidates from the specialist and $16$ from the generalist, for a total of $32$ candidates per question.
Within each source, samples are drawn with a fixed temperature profile and multiple prompt routes; route and decoding metadata are attached to every candidate as provenance signals consumed by the selector.

\subsection{Evaluation Protocol Details}
\label{sec:appendix-evaluation}

All results are reported under each benchmark's official evaluation protocol.\footnote{We use the public scripts distributed by the benchmark organizers without modifying correctness criteria.}
\bird{} adopts the stricter criterion: beyond matching the set of returned rows, the element order \emph{within} each row (i.e., column order) must also match the gold result.
\spider{} ignores column order and only requires value equivalence on each row.
The BIRD difficulty breakdown (simple / moderate / challenging) follows the official BIRD release.

\subsection{Backbone-Matched Baseline Configurations}
\label{sec:appendix-baselines}

The ``backbone-matched'' baselines in Table~\ref{tab:main-results} all share the same scaffold as \system{} (schema linking, prompt routes, candidate budget) and differ only in the candidate sources or the selector:

\begin{itemize}[leftmargin=*,itemsep=1pt,topsep=2pt]
    \item \textbf{Qwen3 self-consistency.} Samples $16$ SQL candidates from Qwen3-235B-Instruct-2507~\citep{yang2025qwen3} and resolves them by majority voting only, with no lifecycle and no hybrid selector.
    \item \textbf{Sothis-only pipeline.} Uses only the specialist generator Sothis-32B for all $32$ candidates, then runs the full execution-grounded lifecycle and hybrid selector.
    \item \textbf{Qwen3-only pipeline.} Uses only the generalist Qwen3-235B-Instruct-2507~\citep{yang2025qwen3} for all $32$ candidates, then runs the full lifecycle and hybrid selector.
    \item \textbf{\system{} (full).} Pairs Sothis-32B with the generalist partner ($16{+}16$ candidates) and runs the full lifecycle and hybrid selector.
\end{itemize}

The prompted single-model rows at the top of Table~\ref{tab:main-results} (Qwen3-Coder, Qwen3-235B, Gemini-3.1 Pro, Sothis-32B) use the same linked-schema prompt under 5-shot prompting but remove the lifecycle and selector entirely, reporting the single-pass EX of each backbone.

\subsection{Schema Context Construction}
\label{sec:schema-details}

We formalize the question-specific schema context introduced in
\S\ref{sec:candidate-construction}.
Given a question $q$ and a target database $d$, the runtime always loads an
offline schema prompt
\begin{equation}
s=\Phi(q,d),
\end{equation}
where $s$ is a prompt-ready schema description (table DDLs, column names,
foreign-key edges, inline per-column comments, and short per-column examples).
Optionally, the runtime may also inject a question-level value-hint block
\begin{equation}
h=\Psi(q,d),
\end{equation}
when such hints are provided by an external cache.
The main runs reported in this paper always use the offline schema prompt $s$;
value hints $h$ are supported by the implementation but are optional rather than
assumed. When available, they are forwarded together with the question and
evidence as structured grounding context $(q,e,s,h)$.\footnote{The offline
schema cache is keyed by question identifier. Optional value hints are keyed by
question and database identifier so that all generators see the same side
information when hints are enabled.}
This design keeps the online pipeline lightweight and makes candidate generation
reproducible across runs.

\subsubsection{Schema Representation Format}
\label{sec:schema-format}

We adopt an \emph{enriched DDL} representation that augments the standard
\texttt{CREATE TABLE} skeleton with two additional information channels:
(i)~\textbf{column comments} that disambiguate cryptic or abbreviated column
names, and (ii)~\textbf{question-conditioned example values} that ground the
model's understanding of the data distribution.
Unless otherwise specified, this subsection describes the default enriched
schema context (DES) used by our main pipeline.
Listing~\ref{lst:schema-full} shows a representative fragment of our enriched
schema format; Listing~\ref{lst:schema-plain} shows the same tables under a
plain DDL baseline~\cite{wang2025agentar} that retains only structural metadata.

\begin{lstlisting}[caption={Enriched schema format used by \system (excerpt from the \texttt{financial} database).}, label=lst:schema-full]
CREATE TABLE account (
    account_id integer, -- example: [1, 2]
    district_id integer, -- location of branch, example: [18, 1]
    frequency text, -- example: ['POPLATEK PO OBRATU', 'POPLATEK MESICNE']
    `date` date, -- example: ['1995-03-24', '1993-02-26']
    PRIMARY KEY (account_id),
    CONSTRAINT fk_account_district_id FOREIGN KEY (district_id) REFERENCES district (district_id)
);

CREATE TABLE district (
    district_id integer, -- location of branch, example: [1, 2]
    A2 text, -- district_name, example: ['Hl.m. Praha', 'Benesov']
    A3 text, -- region, example: ['east Bohemia', 'Prague']
    A4 text, -- number of inhabitants, example: ['1204953', '88884']
    A5 text, -- no. of municipalities with inhabitants < 499, example: ['0', '80']
    PRIMARY KEY (district_id)
);
\end{lstlisting}

\begin{lstlisting}[caption={Plain DDL schema baseline (same tables, no comments or examples).}, label=lst:schema-plain]
CREATE TABLE account (
    account_id integer,
    district_id integer,
    frequency text,
    `date` date,
    PRIMARY KEY (account_id),
    CONSTRAINT fk_account_district_id FOREIGN KEY (district_id) REFERENCES district (district_id)
);

CREATE TABLE district (
    district_id integer,
    A2 text,
    A3 text,
    A4 text,
    A5 text,
    PRIMARY KEY (district_id)
);
\end{lstlisting}

\noindent \textbf{Design rationale.}
The enriched format addresses three failure modes commonly observed when
LLMs operate on plain DDL alone:

\begin{enumerate}[leftmargin=*,itemsep=2pt,topsep=3pt]
\item \textbf{Ambiguous column semantics.}
Columns such as \texttt{A2}--\texttt{A5} in the \texttt{district} table
carry no self-explanatory meaning.
The comment annotation (e.g., \texttt{-- district\_name}) resolves this
ambiguity without requiring the model to infer semantics from data alone.
A comment is emitted only when the human-authored column description
differs from the column name itself; otherwise the comment is suppressed
to avoid redundancy.

\item \textbf{Value-space grounding.}
Example values expose the actual vocabulary stored in the database.
This helps the model produce correct literal predicates in
\texttt{WHERE} clauses without hallucinating non-existent values.
Up to six values per column are shown, drawn from two complementary
sources: (a)~a distinct non-null sample of values from the database,
and (b)~question-relevant values retrieved via BM25 from a prebuilt
database-content index.

\item \textbf{Structural completeness.}
Both formats preserve primary-key and foreign-key constraints verbatim,
ensuring that join-path reasoning is available regardless of the
enrichment level.
Column identifiers containing reserved words or special characters are wrapped
in backticks.
\end{enumerate}

\noindent \textbf{Construction pipeline.}
The schema $s{=}\Phi(q,d)$ is assembled offline by the following steps:
\begin{enumerate}[leftmargin=*,itemsep=1pt,topsep=2pt]
\item Load the database metadata (table names, column names, column
  comments, column types, primary keys, foreign keys) from the
  schema catalogue.
\item For each column, sample up to $k$ distinct non-null values from
  the SQLite database (truncated to 40 characters for long strings).
\item Tokenize the question $q$ (augmented with any external knowledge
  $e$) into $n$-grams ($n{\leq}8$) and retrieve matching database
  cell values via a Lucene-based BM25 index built over all column
  contents.
\item Merge sampled and retrieved values per column (deduplicated,
  capped at six), and emit the enriched DDL with inline comments and
  example annotations.
\end{enumerate}
When optional value hints are enabled, the hint set $h{=}\Psi(q,d)$ is the
subset of retrieved values whose substring-match score against the question
exceeds $0.85$, surfaced separately in the prompt as explicit value-link hints.

\subsubsection{Schema Variant Ablations}
\label{sec:schema-variant-ablations}

In addition to DES, we
construct eight controlled schema variants to isolate the effects of schema
scope and metadata richness. All eight variants are serialized in the same
enriched-DDL template so that the ablation changes the schema content rather
than the prompt format.

\noindent \textbf{Scope.}
We consider two schema scopes:
\begin{enumerate}[leftmargin=*,itemsep=2pt,topsep=3pt]
\item \textbf{Full.}
The full variant keeps all tables and columns in the target database together
with their primary-key and foreign-key constraints.
\item \textbf{Pruned.}
The pruned variant first performs question-conditioned schema linking and then
retains only the linked tables and columns. This is an instance-specific schema
selection step rather than generic prompt truncation.
\end{enumerate}

\noindent \textbf{Metadata levels.}
Under each scope, we instantiate four metadata configurations:
\begin{enumerate}[leftmargin=*,itemsep=2pt,topsep=3pt]
\item \textbf{Plain.}
This is the minimal variant in the ablation family. It keeps only the basic
schema serialization under our template and does not add question-conditioned
retrieved values, value descriptions, or value-range annotations.
\item \textbf{Retrieved examples.}
This variant augments columns with question-conditioned example values from the
embedding-based value-retrieval pipeline. Concretely, an LLM first extracts a
compact set of salient keywords from the question and evidence. These keywords
are embedded and used to retrieve semantically related cell values from a
per-database value index on a column-wise basis. The retrieved values are then
inserted into the example field of the corresponding columns under a fixed
budget.
\item \textbf{Retrieved examples + value meaning.}
This variant additionally exposes human-authored description fields when they
are available. In our implementation, this bundle may include expanded column
names, short column descriptions, and value descriptions, so the model sees
both retrieved example values and natural-language explanations of what a
column or its values mean.
\item \textbf{Retrieved examples + value meaning + value range.}
This is the richest variant. In addition to retrieved examples and description
fields, it adds coarse value-space constraints. For numeric columns, we provide
minimum/maximum-style range information. For low-cardinality categorical
columns, we expose small valid-value sets.
\end{enumerate}

\noindent \textbf{Relation to the default schema.}
These eight variants should not be conflated with the default
enriched schema context. DES uses
question-conditioned lexical retrieval over database contents via $n$-gram
expansion and BM25 matching, then merges the retrieved values with sampled
column values. By contrast, the ablation variants with
\textit{retrieved examples} use an LLM-guided keyword-extraction stage followed
by embedding-based value retrieval, which changes both the retrieval mechanism
and the resulting example set. Similarly, the \textit{plain} variant in this
ablation family is not the same as the pure DDL baseline in
Listing~\ref{lst:schema-plain}; it is simply the least enriched member within
our unified ablation template.

\subsubsection{Schema Variant Results and Prompt Size}
\label{sec:schema-variant-results}

Table~\ref{tab:schema-variant-full} reports all schema-variant results on the
\bird{} dev set, computed over all $1{,}534$ questions. For Qwen3-Coder, the
strongest controlled variant is \texttt{F+E}, which leads that family on all
three metrics: $71.45$ Final EX, $71.25$ majority-voting EX, and
$78.16$ Oracle@$k$. For Qwen3-235B, DES remains the strongest overall setting
at $71.77$ Final EX, $71.45$ majority-voting EX, and $77.38$ Oracle@$k$,
while the strongest controlled variants are more fragmented:
\texttt{F+E} leads Final EX ($71.58$) and Oracle@$k$ ($77.31$), whereas
\texttt{F+E+M} reaches the best majority-voting score within that family
($71.19$). Across both backbones, adding \textit{retrieved examples}
consistently helps the corresponding \textit{plain} schemas, whereas further
adding \textit{value meaning} and \textit{value range} does not produce stable
gains and often slightly degrades the final metric. From the joint perspective
of stability across backbones and absolute effectiveness, DES remains the most
reliable default choice: it is best on Qwen3-235B and stays close to the best
controlled variant on Qwen3-Coder without requiring the richer retrieval-based
variant stack.

\begin{table*}[t]
    \centering
    \caption{Schema-variant results on \bird{} dev (\%). DES: default enriched schema. \texttt{F}: full, \texttt{P}: pruned, \texttt{E}: retrieved examples, \texttt{M}: value meaning, \texttt{R}: value range.}
    \label{tab:schema-variant-full}
    \small
    \setlength{\tabcolsep}{3pt}
    \renewcommand{\arraystretch}{0.95}
    \begin{tabular}{@{}lcccccc@{}}
        \toprule
        & \multicolumn{3}{c}{Qwen3-Coder} & \multicolumn{3}{c}{Qwen3-235B} \\
        \cmidrule(lr){2-4}\cmidrule(lr){5-7}
        Variant & Final EX $\uparrow$ & Maj $\uparrow$ & Oracle@$k$ $\uparrow$ & Final EX $\uparrow$ & Maj $\uparrow$ & Oracle@$k$ $\uparrow$ \\
        \midrule
        DES & 70.99 & 70.53 & 77.71 & \textbf{71.77} & \textbf{71.45} & \textbf{77.38} \\
        \texttt{F} & 69.82 & 69.49 & 76.79 & 70.93 & 70.73 & 76.66 \\
        \texttt{F+E} & \textbf{71.45} & \textbf{71.25} & \textbf{78.16} & 71.58 & 71.06 & 77.31 \\
        \texttt{F+E+M} & 71.06 & 71.12 & 77.05 & 71.25 & 71.19 & 77.25 \\
        \texttt{F+E+M+R} & 70.99 & 70.73 & 77.64 & 70.66 & 70.80 & 76.92 \\
        \texttt{P} & 70.14 & 69.17 & 75.68 & 71.12 & 71.06 & 75.75 \\
        \texttt{P+E} & 71.06 & 70.80 & 76.92 & 71.25 & 71.12 & 76.01 \\
        \texttt{P+E+M} & 70.34 & 70.14 & 76.47 & 70.80 & 70.53 & 75.81 \\
        \texttt{P+E+M+R} & 70.86 & 70.60 & 76.47 & 71.06 & 70.60 & 76.14 \\
        \bottomrule
    \end{tabular}
\end{table*}

Table~\ref{tab:schema-variant-size} complements the accuracy comparison with
prompt-side schema length. The full variants are expensive, ranging from
$2.9$k to $4.5$k tokens on average, while the pruned variants fall between
roughly $326$ and $481$ tokens. In other words, pruning reduces the average
schema length by nearly an order of magnitude. However, this compression does
not translate into a consistent EX improvement in Table~\ref{tab:schema-variant-full}.
The default enriched schema context occupies a middle point at roughly
$1.95$k tokens on average while remaining competitive with, and in one case
stronger than, the richer full variants with embedding-retrieved examples.
Taken together, the results indicate that better question-conditioned value
grounding matters more than simply adding more schema-side metadata or
aggressively minimizing prompt length, and they further support retaining DES
as the primary setting in the main system.

\begin{table}[t]
    \centering
    \caption{Prompt-side schema size of the same variants, measured in Qwen3-Coder tokens.}
    \label{tab:schema-variant-size}
    \small
    \setlength{\tabcolsep}{3pt}
    \renewcommand{\arraystretch}{0.95}
    \begin{tabular}{@{}lrrrrrrr@{}}
        \toprule
        Variant & Avg. & Min. & Max. & p50 & p90 & p95 & p99 \\
        \midrule
        DES & 1951 & 383 & 4654 & 2051 & 2970 & 4641 & 4654 \\
        \texttt{F} & 2858 & 519 & 6172 & 3174 & 5188 & 6172 & 6172 \\
        \texttt{F+E} & 3148 & 649 & 6449 & 3216 & 5855 & 6417 & 6433 \\
        \texttt{F+E+M} & 3821 & 672 & 7765 & 3419 & 7128 & 7733 & 7749 \\
        \texttt{F+E+M+R} & 4521 & 837 & 10348 & 3982 & 7859 & 10316 & 10332 \\
        \texttt{P} & 326 & 30 & 2444 & 287 & 521 & 684 & 1401 \\
        \texttt{P+E} & 361 & 30 & 2610 & 318 & 603 & 744 & 1487 \\
        \texttt{P+E+M} & 411 & 49 & 2730 & 336 & 689 & 842 & 1569 \\
        \texttt{P+E+M+R} & 481 & 52 & 3561 & 432 & 789 & 941 & 1954 \\
        \bottomrule
    \end{tabular}
\end{table}

\section{L1: Specialist Training and Complementary Candidate Construction}
\label{sec:training-details}

This section expands L1 in \S\ref{sec:training} with the training data
composition, the combined training objective (DAPO + threshold-gated
entropy), the training-time execution sandbox, and the two-stage DAPO
configuration used for the \system specialist generator.
We train the specialist under reinforcement learning with verifiable rewards
(RLVR): the environment supplies reliable supervision by executing sampled
SQL programs, but the reward is observed only after a complete response is
produced.
This terminal-only signal makes policy collapse a practical failure mode.
As the model overfits easy prompts, its response distribution can become too
sharp, triggering entropy collapse and reducing the sampling diversity needed
to cross discontinuous execution-reward boundaries on harder tasks. In
group-relative training, this appears as all-correct or all-incorrect rollout
groups, where the in-group advantage is zero and the prompt contributes no
useful gradient.
Our recipe therefore balances exploration and exploitation from two sides:
we reshape the training data to reduce zero-variance prompts, and we use a
dynamic-sampling DAPO schedule with entropy control to keep policy updates
well conditioned.

\subsection{Training Data}
\label{sec:training-data}
The specialist is trained on a union of three Text-to-SQL sources converted
into a single prompt format: the \bird training split, the \spider training
split, and a sample from SynSQL~\citep{li2025omnisql}. Each example is
serialized with the default enriched schema context (DES) described in
\S\ref{sec:schema-format}, so the training prompt exposes table structure,
column descriptions, and representative values in the same format used at
inference time.

\noindent \textbf{Data construction and cleaning.}
We apply a two-stage cleaning pipeline before RL. First, the execution
environment drops examples whose gold SQL fails to execute, times out, or
returns an empty result on the paired database. This avoids introducing
zero-reward targets that would inflate the number of zero-variance prompts
during RL. Second, we use a multi-model voting filter with DeepSeek-v3,
GPT-5, and Qwen3-235B to check semantic consistency from multiple angles,
including schema selection, join structure, and predicate constraints. Examples
that fail this cross-model consistency check are removed from the training
pool, reducing noisy supervision before policy optimization. The resulting
examples, together with the SynSQL sample, are consolidated into a single
DAPO training set.
Before DAPO training, we reserve 500 examples from the training sources as a validation split for monitoring training dynamics; these examples are excluded from policy updates.
Table~\ref{tab:training-data-mixture}
summarizes the final composition.

\begin{table}[t]
\centering
\small
\begin{tabular}{lrr}
\toprule
\textbf{Source} & \textbf{Rows} & \textbf{Share} \\
\midrule
SynSQL sample   & 20{,}484 & 75.4\% \\
\bird train     &  4{,}726 & 17.4\% \\
\spider train   &  1{,}972 &  7.3\% \\
\midrule
Total (train)   & 27{,}182 & 100\%  \\
Validation (train sources) &      500 & --     \\
\bottomrule
\end{tabular}
\caption{Specialist training data composition after execution filtering and
multi-model semantic-consistency filtering. SynSQL is subsampled from the
public release. The validation split is reserved from the training sources before DAPO optimization.}
\label{tab:training-data-mixture}
\end{table}

\begin{figure*}[t]
    \centering
    \includegraphics[width=0.98\textwidth]{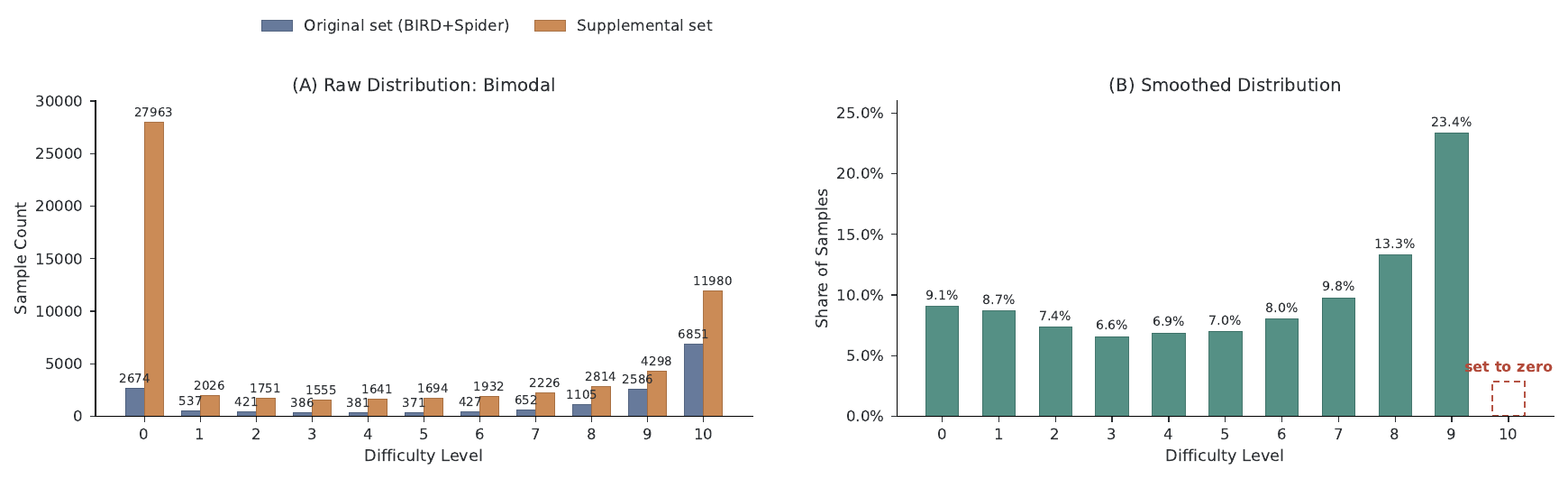}
    \caption{Difficulty distribution before and after smoothing. The x-axis is
    the number of correct baseline rollouts out of 10. Smoothing redistributes
    training mass over informative buckets and sets the trivially solved
    $10/10$ bucket to zero.}
    \label{fig:difficulty-distribution}
\end{figure*}

\noindent \textbf{Difficulty smoothing.}
The difficulty distribution of the raw training pool directly affects the
quality of policy gradients. Very easy prompts mostly reinforce behavior that
the model can already exploit, providing limited training signal.
Moderately hard prompts are the most useful for RL because they still admit
occasional successes while forcing the policy to explore beyond its current
high-probability outputs. We estimate prompt
difficulty by sampling a baseline model for $10$ independent rollouts and
grouping examples by the number of correct rollouts: $0/10$ denotes prompts
the baseline never solves, while $10/10$ denotes prompts solved in every
rollout. The raw pool is strongly imbalanced, with heavy mass at the easy
end and a long tail of hard prompts (Fig.~\ref{fig:difficulty-distribution}).
We reshape the data distribution by smoothing overrepresented
pass-rate buckets and removing the trivially solved $10/10$ bucket, so RL
batches concentrate on prompts with informative rollout variation. The
$10/10$ bin is retained in the figure to show that these prompts are
intentionally emptied by the smoothing procedure rather than omitted from the
axis.

\subsection{DAPO Objective, Dynamic Sampling, and Threshold-Gated Entropy}
\label{sec:training-objective}
The total training objective combines the DAPO policy loss with a
threshold-gated entropy regularizer:
\begin{equation}
\mathcal{L}_{\mathrm{total}}(\theta)
= \mathcal{L}_{\mathrm{DAPO}}(\theta)
+ w_{\mathrm{ent}}\,\mathcal{L}_{\mathrm{ent}}(\theta).
\end{equation}

\noindent \textbf{Reward.}
Each sampled SQL receives a scalar execution reward from the training-time
sandbox of \S\ref{sec:training-sandbox}, taking one of three values:
$R{=}1$ when the SQL is executable and its row set matches the gold,
$R{=}0.1$ when it is executable but the rows do not match, and $R{=}0$
when it is unparsable or triggers a runtime or timeout error.
The small partial-credit value of $0.1$ for executable-but-wrong rollouts
is important in the zero-variance-prompt regime: when all $n$ rollouts
fail to match the gold, the partial-credit term nonetheless separates
``plausibly executable'' rollouts from broken ones, keeping the group
advantage non-degenerate and producing a usable gradient that a pure
$\{0,1\}$ reward would have lost.
An additional soft penalty on overlong generations is applied using a
length buffer of $1024$ tokens and penalty factor $1.0$.

\noindent \textbf{GRPO advantage and dynamic sampling.}
For prompt $x$ with $n$ sampled responses $\{y_i\}_{i=1}^n$ from
$\pi_\theta$, GRPO normalizes the scalar reward within the group:
\begin{equation}
A_{i} = \frac{R(x,y_i) - \mu}{\sigma + \epsilon},
\label{eq:grpo-adv}
\end{equation}
where $\mu$ and $\sigma$ are the in-group mean and standard deviation of
$R(x,y_k)$, and $A_i$ is broadcast to every token in~$y_i$.
Let $\rho_{i,t}(\theta){=}\pi_\theta(y_{i,t}{\mid}
x,y_{i,<t}){/}\pi_{\theta_{\text{old}}}(y_{i,t}{\mid} x,y_{i,<t})$
denote the token-level importance ratio.
Policy optimization uses the per-token asymmetric-clipped
surrogate~\citep{yu2026dapo}:
\begin{equation}
\mathcal{L}_{\mathrm{DAPO}}(\theta)
= \frac{\mathbb{1}[\mathrm{Var}_i R(x,y_i){>}0]}{n}
\sum_{i=1}^{n}\ell_i(\theta),
\end{equation}
\begin{align}
\ell_i(\theta) = \min\!\Bigl(
  &\rho_{i,t}\,A_i,\notag\\
  &\mathrm{clip}(\rho_{i,t},\,1{-}\epsilon_l,\,1{+}\epsilon_h)\,A_i
\Bigr).
\end{align}
The leading indicator implements \emph{dynamic sampling}.
Prompts whose rollouts produce all-correct or all-incorrect groups are
discarded from the gradient update, and up to $e_{max}$ extra rollout batches
are drawn until a training batch of non-degenerate prompts is accumulated.
This rejection-sampling loop is especially important under sparse execution
feedback: it adaptively spends additional inference on prompts whose initial
rollouts do not provide relative advantage, so the optimizer sees batches with
a higher signal-to-noise ratio.
We use $\epsilon_l{=}0.2$ and $\epsilon_h{=}0.28$, and disable KL entirely.

\noindent \textbf{Threshold-gated entropy regularizer.}
Beyond the standard DAPO components above, we add a simple but effective
gate against entropy collapse.
Let $H(\pi_\theta)=-\!\sum_a \pi_\theta(a\mid s_t)\log\pi_\theta(a\mid s_t)$
be the average policy entropy over the current rollout, and let
$\mathcal{L}_{\mathrm{ent}}(\theta)$ denote its empirical estimate:
\begin{equation}
\mathcal{L}_{\mathrm{ent}}(\theta)
= -\frac{1}{T}\sum_{t=1}^{T}\!
\sum_{a\in\mathcal{V}}\!
\pi_\theta(a\!\mid\!s_t)\log\pi_\theta(a\!\mid\!s_t).
\end{equation}
The gating coefficient is a step function of $H(\pi_\theta)$:
\begin{equation}
w_{\mathrm{ent}} = \beta \cdot \mathbf{1}\!\bigl[H(\pi_\theta) < H_{\mathrm{target}}\bigr],
\end{equation}
This means the entropy bonus pays zero regularization tax whenever the policy
already maintains sufficient diversity, and only injects a fixed gradient
$\beta$ once the rollouts start collapsing.
Unlike a constant entropy coefficient, this ``activate-on-collapse'' schedule
preserves the execution-reward signal during normal optimization and intervenes
only at the boundary of entropy failure.
The target entropy is stage-specific (Table~\ref{tab:training-hparams}), and
we set $\beta{=}5{\times}10^{-3}$. In practice $\beta$ is warmed up with step
$\Delta{=}10^{-4}$ to avoid abrupt gradient spikes when the gate first opens.

\subsection{Training-Time Execution Sandbox}
\label{sec:training-sandbox}
Because the reward requires running every rollout against a real database at
every optimization step, we wrap SQL execution in a reproducible training-time
sandbox. Each rollout and its gold SQL are executed on the target SQLite
database through a read-only connection, their result sets are compared as
unordered sets, and the three-level reward rule above is applied. A
$60$-second wall-clock budget is enforced per query: timeouts and
runtime errors are both mapped to reward $0$, while executable-but-wrong
queries receive the partial credit $0.1$.

\subsection{Specialist DAPO Configuration}
Table~\ref{tab:training-hparams} summarizes the specialist DAPO
configuration. Our training recipe follows a two-stage schedule sharing the
same objective: an exploration stage with a wide rollout ($n{=}32$,
temperature $1.2$, $H_{\mathrm{target}}{=}0.30$) that helps collect sparse
successful trajectories on hard prompts, and an exploitation stage that
resumes from the exploration checkpoint and anneals the sampling width
($n{=}16$, temperature $0.9$, $H_{\mathrm{target}}{=}0.15$). The first stage
uses broad sampling and a higher entropy target to prevent early
homogenization as previously challenging prompts become easy. The second
stage applies a simulated-annealing style constraint: it lowers the sampling
temperature, reduces the rollout width, and removes the entropy target so that
optimization shifts from broad state-space exploration toward reinforcing
verified high-confidence solutions with less rollout variance.

\begin{figure}[t]
    \centering
    \includegraphics[width=\columnwidth]{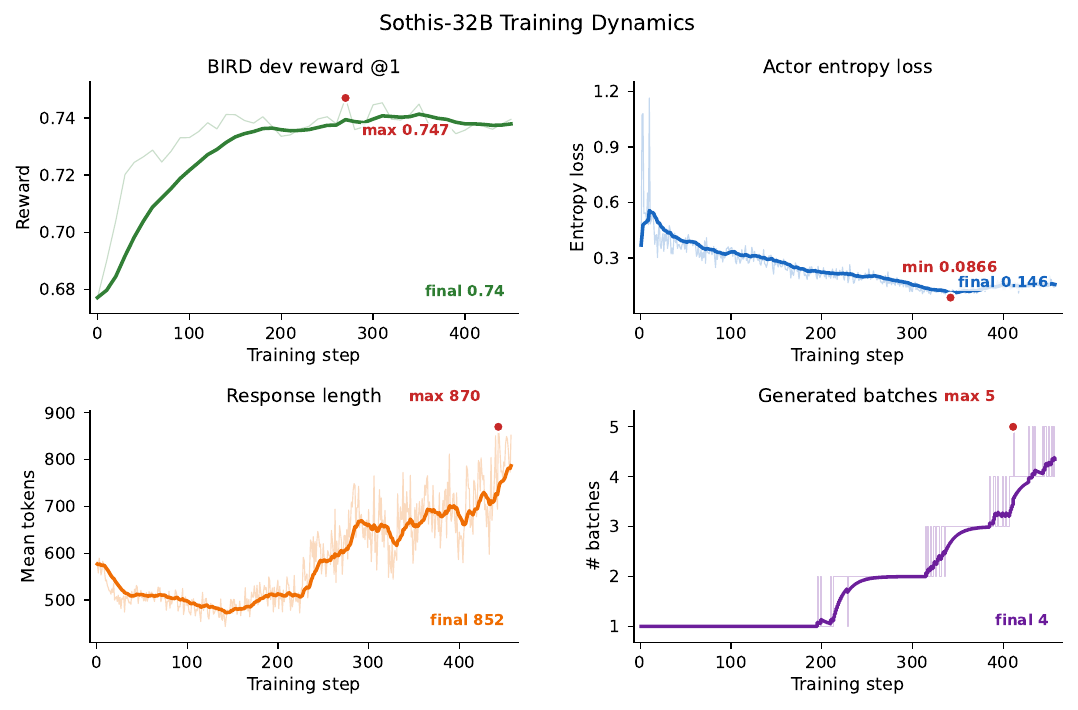}
    \caption{Sothis-32B training dynamics under the two-stage DAPO recipe.
    Validation reward improves and stabilizes; entropy loss decays without
    immediate collapse; response length first contracts and later recovers as
    SQL completions become more detailed; generated batches increase when
    dynamic sampling needs extra rollouts to construct non-degenerate updates.}
    \label{fig:training-curves-32b}
\end{figure}

Figure~\ref{fig:training-curves-32b} summarizes these dynamics. The reward
curve stabilizes near its final value rather than peaking and immediately
collapsing. The entropy-loss curve decreases as the policy becomes more
confident, while the generated-batch curve rises later in training, indicating
that dynamic sampling increasingly has to search for rollout groups with
non-zero relative advantage as the model solves more prompts consistently.

\begin{table}[t]
\centering
\small
\begin{tabular}{lcc}
\toprule
& \textbf{Exploration} & \textbf{Exploitation} \\
\midrule
Base model             & \multicolumn{2}{c}{OmniSQL-32B} \\
Adv.\ estimator        & \multicolumn{2}{c}{GRPO} \\
Rollout $n$            & 32     & 16    \\
Rollout temperature    & 1.2    & 0.9   \\
Clip ratio low/high    & \multicolumn{2}{c}{0.2 / 0.28} \\
KL regularization      & \multicolumn{2}{c}{disabled} \\
Entropy target $H_{\mathrm{target}}$ & $0.30$ & $0.15$ \\
Entropy coef.\ $\beta$  & \multicolumn{2}{c}{$5{\times}10^{-3}$} \\
Entropy warm-up $\Delta$ & \multicolumn{2}{c}{$10^{-4}$} \\
Max prompt / response  & \multicolumn{2}{c}{6000 / 5120} \\
Overlong buffer        & \multicolumn{2}{c}{1024 tokens, factor 1.0} \\
Dyn.\ sampling retries & \multicolumn{2}{c}{up to 10 batches} \\
Train / gen / mini bsz & \multicolumn{2}{c}{128 / 384 / 128} \\
Learning rate          & \multicolumn{2}{c}{$1{\times}10^{-5}$} \\
LR warmup steps        & \multicolumn{2}{c}{10} \\
Weight decay           & \multicolumn{2}{c}{0.1} \\
Gradient clip          & \multicolumn{2}{c}{1.0} \\
Epochs                 & 20     & 10    \\
Hardware               & \multicolumn{2}{c}{4\,$\times$\,8 GPUs} \\
\bottomrule
\end{tabular}
\caption{Two-stage specialist training configuration.
Exploitation resumes from the exploration checkpoint and anneals the sampling
temperature and rollout width while preserving dynamic sampling.}
\label{tab:training-hparams}
\end{table}

\subsection{Critic Reward under Difficulty-Smoothed Data}
\label{sec:training-critic}

\begin{figure}[t]
    \centering
    \includegraphics[width=\columnwidth]{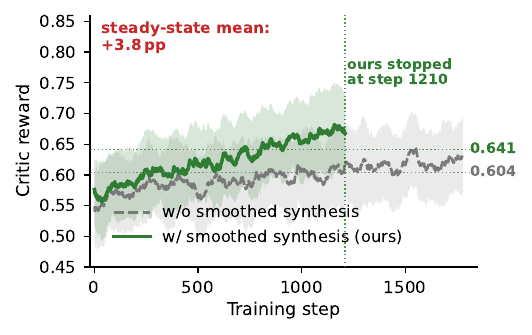}
    \caption{Critic reward during specialist RL.}
    \label{fig:training-curves-critic}
\end{figure}

Figure~\ref{fig:training-curves-critic} reports the critic reward during
specialist RL under the same two configurations compared in
Fig.~\ref{fig:training-curves}.
With difficulty-smoothed data synthesis the critic reward reaches a
higher steady-state mean ($0.641$ vs.\ $0.604$, $+3.8$\,pp) and maintains
a visibly larger variance band throughout training, consistent with a
non-collapsing policy that continues to explore novel trajectories after
the baseline has plateaued.
We stop training at step $1{,}210$ for the smoothed run because the
critic curve has entered a stable regime.

\section{L2: Execution-Grounded Candidate Lifecycle}
\label{sec:lifecycle-details}

Algorithm~\ref{alg:lifecycle} gives the full lifecycle pseudocode before the block-level details.

\begin{algorithm}[!htbp]
\caption{Execution-Grounded Candidate Lifecycle}
\label{alg:lifecycle}
\begin{algorithmic}[1]
\Require input $x=(q,e,d)$, linked context $(s,h)$, generators $G_{\text{sp}},G_{\text{ge}}$
\Ensure pool $\mathcal{C}^{(*)}$
\State $\mathcal{C}^{(0)} \gets G_{\text{sp}}(q,e,s,h) \cup G_{\text{ge}}(q,e,s,h)$
  \Comment{candidate construction}
\State $(\mathcal{C}^{(1)}, \mathcal{R}_{\text{phase2}}) \gets \mathsf{Validate}(\mathcal{C}^{(0)})$
  \Comment{pre-execution screening}
\ForAll{$c \in \mathcal{C}^{(1)}$}
    \State $\delta(c) \gets \mathsf{Exec}(c)$
    \Comment{$\delta\!\in\!\{\texttt{clean},\texttt{runtime},\texttt{timeout},\texttt{empty}\}$}
\EndFor
\ForAll{$c$ with $\delta(c)\neq\texttt{clean}$}
    \If{$\delta(c)=\texttt{empty}$}
        \State $c' \gets R_{\mathrm{op}}(c;\,\texttt{operators})$
        \Comment{high-precision empty-result operators first}
        \If{$c'$ still returns empty}
            \State $c' \gets R_{\mathrm{empty\_llm}}(c;\,q,e,s,h)$
            \Comment{diagnosis-conditioned Stage-2 fallback}
        \EndIf
    \Else
        \State $c' \gets R_{\delta(c)}(c;\,q,e,s,h)$
        \Comment{runtime/timeout/validation repair}
    \EndIf
    \If{$c' \neq \varnothing$}
        \State $\delta(c') \gets \mathsf{Exec}(c')$
        \Comment{immediate re-execution}
        \State replace $c$ with $c'$ if $\delta(c')=\texttt{clean}$, else mark $c$ unresolved
    \EndIf
\EndFor
\State $\mathcal{C}^{(*)} \gets \mathsf{NormalizeProjection}(\mathsf{Consolidate}(\mathcal{C}^{(1)}))$
\If{$\mathcal{C}^{(*)}=\varnothing$}
    \State build condensed negative feedback $(\mathcal{B},\rho)$ from unresolved and rule-filtered SQLs
    \State $\mathcal{C}^{(*)} \gets G_{\mathrm{rec}}(q,e,s;\mathcal{B},\rho)$
    \Comment{zero-candidate recovery}
\EndIf
\State \Return $\mathcal{C}^{(*)}$
\end{algorithmic}
\end{algorithm}

This section expands \S\ref{sec:lifecycle} of the main paper and
follows the three blocks of the Execution-Grounded Lifecycle panel in
Figure~\ref{fig:pipeline}: \textbf{(A)~Pre-execution Screening},
\textbf{(B)~Execution-Grounded Diagnosis}, and
\textbf{(C)~Failure-Aware Pool Update}. Block~A applies deterministic
sanitization and lightweight syntax/schema validation, removing malformed
candidates from the pool before execution. Block~B executes every
surviving candidate and tags each with one of four diagnostic labels
\{\texttt{clean},\texttt{runtime},\texttt{timeout},\texttt{empty}\}.
Block~C dispatches each non-\texttt{clean} candidate to a dedicated
type-specific repair path, re-executes accepted rewrites immediately,
normalizes projections deterministically, and regenerates only when the
pool is fully exhausted.

\subsection{Block~A: Pre-execution Screening}
\label{sec:lifecycle-block-a}

Before any execution or LLM correction, candidates are sanitized by a
short set of deterministic preprocessors targeted at common
LLM-generation artifacts: stripping \texttt{--}/\texttt{\#} comment
prefixes that precede the actual SQL, inserting a missing space after
the \texttt{SELECT} keyword when the LLM emits \texttt{SELECT*}, and
rejecting outputs whose first token is not a valid SQL leading keyword
(\texttt{SELECT}, \texttt{WITH}, \texttt{INSERT}, \texttt{UPDATE},
\texttt{DELETE}, \texttt{CREATE}, \texttt{DROP}).
The sanitized SQL is then passed through dual validation.
The syntax validator is a lightweight rule-based checker rather than a full SQL
parser: it verifies a legal leading keyword, required \texttt{FROM} presence
for \texttt{SELECT}, balanced parentheses outside string literals, balanced
quotes, and a few common comma-related artifacts.
The schema validator then combines simple table-name extraction with an
\texttt{EXPLAIN}-based check that catches missing tables, missing columns,
ambiguous columns, and other SQLite-side schema mismatches.
Candidates failing either check are removed from the pool in Phase~2,
with the specific failure subtype attached for diagnostics and later summary
statistics.

\subsection{Block~B: Execution-Grounded Diagnosis}
\label{sec:lifecycle-block-b}

Block~B executes every candidate that survives pre-execution screening and
attaches a diagnostic label used by the downstream repair dispatcher.
A candidate is marked \texttt{clean} when execution succeeds and returns a
non-empty result, \texttt{empty} when execution succeeds but returns zero rows,
\texttt{runtime} when SQLite raises an engine-level exception, and
\texttt{timeout} when execution exceeds the wall-clock budget.
The label is retained together with the original SQL, execution feedback, and
source-route metadata, so Block~C can choose a type-specific repair path rather
than treating all failed candidates as interchangeable correction requests.

\subsection{Block~C: Failure-Aware Repair}
\label{sec:lifecycle-block-c}

Each non-\texttt{clean} candidate in the active post-Phase2 pool is routed to
one of four repair paths.
All four paths share a common prompt skeleton over question, evidence, schema,
current SQL, failure type, and failure feedback.
They differ only in the concrete feedback and in
\texttt{type\_specific\_instructions}.
Every repaired SQL is re-executed immediately.
The repaired candidate replaces the original only if re-execution
is \texttt{clean} (or \texttt{clean} with a non-empty result for
\texttt{empty} repairs); otherwise the original is moved
out of the pool and consolidation proceeds.

\noindent \textbf{Validation repair.}
The implementation contains a validation-repair path for candidates whose
current state is still marked invalid at repair time. In the normal runtime
used in our main experiments, however, Phase~2 removes validation failures
from the pool before Phase~4 begins, so repair budget is spent almost
entirely on execution-surviving failures. When validation repair is invoked,
the feedback slot lists the specific syntax or schema violations surfaced by
Block~A (e.g., unknown column, missing table, mismatched alias), and the
type-specific instructions emphasize repairing the structural error while
preserving the original query intent.

\noindent \textbf{Runtime repair.}
The feedback slot carries the SQLite engine's error string together
with any residual validation context. The instructions direct the LLM
to address the exact engine-reported root cause (type mismatch, invalid
function call, ambiguous column), not to restructure the whole query.

\noindent \textbf{Timeout repair.}
The feedback slot records the execution elapsed time against the wall
budget. The instructions prioritize eliminating missing \texttt{JOIN}
conditions, Cartesian products, redundant nested subqueries, and other
expensive structural patterns while keeping the semantic intent intact.

\noindent \textbf{Empty-result repair: two-stage cascade.}
Empty-result is the most load-bearing failure class because, unlike
runtime or timeout failures, the query is formally well-formed. The
failure is semantic rather than structural.  We therefore apply the
two-stage cascade shown inside Block~C of Figure~\ref{fig:pipeline}:
front-load cheap, high-precision structural edits and only escalate to
an LLM when structural edits do not apply.

\emph{Stage~1: high-precision operator catalogue.}
Before any LLM call, the candidate is passed through a catalogue of
six deterministic operators, each targeting a specific empty-result
mechanism observed repeatedly in the training split:
\begin{itemize}[leftmargin=*,itemsep=1pt,topsep=2pt]
\item \textbf{\texttt{literal\_binding}}: matches empty results caused
by case, prefix, or substring mismatch between a literal in the
\texttt{WHERE} clause and the actual stored value, and rewrites the
predicate with a probed alternative literal.
\item \textbf{\texttt{literal\_rebind}}: similar to
\texttt{literal\_binding} but probes the database for candidate
replacement values and picks the closest match under a
database-backed similarity check.
\item \textbf{\texttt{where\_predicate}}: rewrites over-restrictive
\texttt{WHERE} shapes such as a chain of equality predicates that
should have been a single \texttt{LIKE}, or a nested subquery that
should have been a direct \texttt{JOIN}.
\item \textbf{\texttt{translation\_semantics}}: handles
natural-language words with multiple SQL semantics
(e.g., ``average'' $\to$ \texttt{AVG} vs.\
\texttt{SUM/COUNT}), replacing the wrong aggregate with the
intent-consistent one.
\item \textbf{\texttt{date\_anchor\_join}}: catches empty results from
date-range predicates on the wrong table in a multi-table schema, and
rewrites the predicate onto the joined table along the identified
foreign key.
\item \textbf{\texttt{aggregate\_scope}}: corrects \texttt{GROUP~BY} or
\texttt{HAVING} scope mismatches, for example moving a predicate from
\texttt{WHERE} to \texttt{HAVING} when the column is an aggregate.
\end{itemize}
Each operator independently emits zero or more candidate repairs, and
the first repair whose re-execution returns a non-empty result is
accepted. If no operator produces a non-empty result, the candidate
escalates to Stage~2.

\emph{Stage~2: error-bank LLM fallback.}
Stage~2 delegates to an iterative error-bank repair loop. The first
round probes the database to produce a structured diagnosis (which
literal value is missing, which join path is cold, which aggregate
scope is wrong) and feeds the diagnosis to a probe-guided repair
prompt. If the repaired SQL still returns empty, the loop issues a
baseline rewrite prompt that instructs the LLM to draft a completely
new SQL for the question while treating the previous attempt as a
negative example. We cap the loop at two rounds\footnote{The cap is a wall-clock and cost guard rather than a modeling assumption. In pilot runs, additional rounds mostly repeated the same value-binding mistakes.}
so that Stage~2
remains bounded in LLM cost.

\subsection{Block~C: Post-Repair Rule Filtering and Pool Recovery}
\label{sec:lifecycle-postrepair}

After re-execution confirmation and consolidation, Block~C applies a small set
of deterministic projection rules that \emph{rewrite} noisy
\texttt{SELECT} lists into more canonical forms.
Two representative rules are:
\begin{itemize}[leftmargin=*,itemsep=1pt,topsep=2pt]
\item \emph{Numbered-column deduplication.} If a projection contains
columns with numeric suffixes alongside the base column, and the question does
not request all variants, the suffixed columns are dropped.
\item \emph{Canonical-bundle reordering.} If a projection contains a standard
column bundle in a non-canonical order, the columns are reordered to the form
typically written by a human SQL author.
\end{itemize}
These rules encode regularities that humans typically would not write
into hand-authored SQL. LLMs, in contrast, sometimes produce
verbatim-complete column lists or unordered bundles when they
over-pattern-match the schema. Applying these adjustments before
selection reduces noise in the pool-level consensus signals of
\S\ref{sec:selection} without discarding otherwise valid queries.

If post-repair consolidation leaves the pool empty for a
question, the system invokes \emph{pool recovery}
(Figure~\ref{fig:pipeline}, Block~C, rightmost box). The regeneration
prompt does not consume the full historical removed set. Instead, it receives
a condensed negative-feedback block built from (i) unresolved or still-invalid
candidates at the end of Phase~4, (ii) candidates filtered by the Phase~5
rules, and (iii) the matched failure tags $\rho$ attached to those removals.
This keeps the feedback specific to the failure modes that actually exhausted
the pool.

\subsection{Additional Lifecycle Diagnostics}
\label{sec:appendix-extra-tables}

\begin{figure*}[thb]
    \centering
    \includegraphics[width=0.95\textwidth]{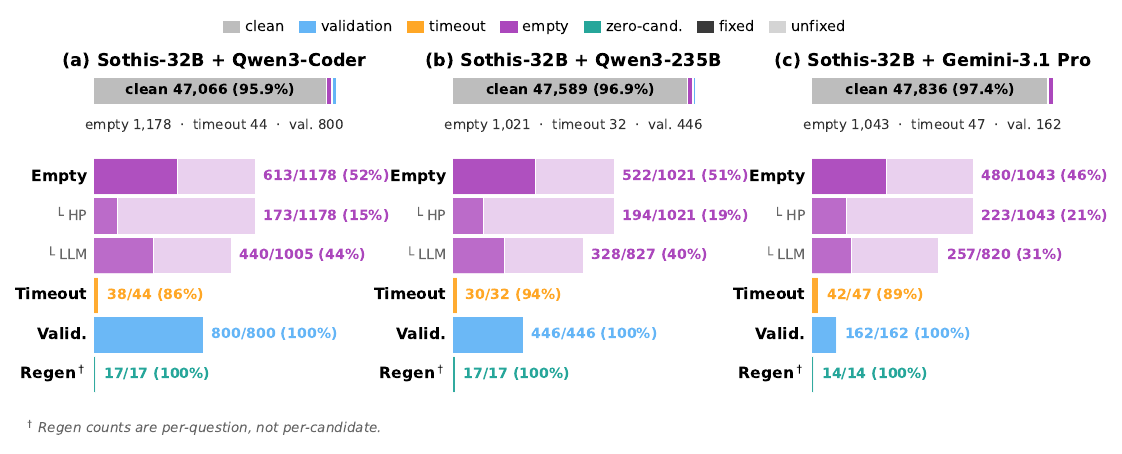}
    \caption{Per-model failure breakdown on \bird{} dev.}
    \label{fig:failure-breakdown-appendix}
\end{figure*}
\noindent \textbf{Per-model failure breakdown.}
Figure~\ref{fig:failure-breakdown-appendix} gives the per-mechanism
candidate-level repair rates broken down by generalist pairing:
Sothis-32B~+~Qwen3-Coder, Sothis-32B~+~Qwen3-235B, and
Sothis-32B~+~Gemini-3.1~Pro.
Across all three pools the qualitative shape of Fig.~4 in the main text
is preserved: validation failures and zero-candidate regeneration each
resolve $100\%$ of their cases, timeouts are resolved at $86$--$94\%$ by
the dual-pool consistency check, and empty-result repair is the hardest
category with total resolution rates in $46$--$52\%$.
The L\,HP row isolates the contribution of the high-precision structural
operators (Stage~1 on empty-result) and the L\,LLM row shows the
additional residue recovered by the LLM fallback (Stage~2).
Empty-result repair depends mildly on the partner generalist because
stronger generalists produce fewer empty-result failures in the first
place (empty counts $1{,}178 \rightarrow 1{,}021 \rightarrow 1{,}043$),
but the relative share recovered by the Stage~1/Stage~2 cascade remains
comparable across pools.

\section{L3: Confidence-Gated Selection}
\label{sec:selection-details}

Algorithm~\ref{alg:selection} gives the full selection procedure before the score definitions and derivations.

\begin{algorithm}[!htbp]
\caption{Confidence-Gated Hybrid Selection}
\label{alg:selection}
\begin{algorithmic}[1]
\Require pool $\mathcal{C}^{(*)}$, input $x=(q,e,d)$, schema $s$
\Ensure final SQL $y$
\State $\mathcal{C}^{+} \gets \{c\in\mathcal{C}^{(*)}\mid\delta(c)=\texttt{clean}\}$
\ForAll{$c \in \mathcal{C}^{+}$}
    \State compute tuple-level consensus $S_{\mathrm{maj}}(c)$
\EndFor
\State Partition $[\mathcal{C}^{+}]_\sigma$ by result signature
\State compute $S_{\mathrm{rm}}^{\mathrm{dedup}}$ over class representatives
\State broadcast $S_{\mathrm{rm}}^{\mathrm{dedup}}$ to class members
\State $S_{\mathrm{coarse}}(c) \gets S_{\mathrm{maj}}(c)+S_{\mathrm{rm}}(c)$
\State compute auxiliary consensus-refine statistic $S_{\mathrm{ref}}(c)$
\State $c^{(1)} \gets \arg\max_c S_{\mathrm{coarse}}(c)$
\State $g^{(1)} \gets$ result-signature class containing $c^{(1)}$
\State $y \gets c^{(1)}$
\Comment{default}
\If{$\mathsf{gate}(x)$ \textbf{holds (\S\ref{sec:selection})}}
    \State extract $\kappa(c)$ for $c$ in the tied top classes
    \State $g^* \gets $ tied-top class with highest cross-source support under $\kappa$
    \If{$\mathrm{supp}(g^*) > \mathrm{supp}(g^{(1)})$}
        \State $y \gets \hat{c}_{g^*}$
        \Comment{structural-support override}
    \EndIf
\EndIf
\State \Return $y$
\end{algorithmic}
\end{algorithm}

This section expands L3 in \S\ref{sec:selection} with the full
derivations that are only summarized in the main text.

\noindent \textbf{Confidence-gate thresholds.}
The cardinality bounds in the confidence gate (\S\ref{sec:selection}) are set to
$K_{\max}{=}6$ and $T_{\max}{=}8$ in all reported runs.
The three score-gap thresholds are all zero, so the selector escalates only
when the merged score, the auxiliary consensus-refine score, and the RM score
all leave the top candidate unresolved.
These values are chosen so that the gate excludes both
\emph{over-fragmented} pools (many tiny result-signature classes with no
dominant candidate, where the structural-support prior becomes unreliable)
and \emph{over-concentrated} pools (one class already taking the majority of
the pool, where $c^{(1)}$ is already the correct default).
In practice the gate fires on roughly 5--10\% of \bird dev questions.\footnote{This rate varies slightly with the generalist partner because candidate pools differ in the number and size of result-signature classes.}

\subsection{Tuple-Level Consensus Derivation}
\label{sec:selection-consensus}

Standard self-consistency for Text-to-SQL treats each candidate's execution
result as an atomic signature and selects the one produced by the most
candidates~\citep{wang2025mac}.
This whole-result voting fails when (i)~no two candidates produce identical
row sets, (ii)~superficial differences such as column ordering or alias
choice create spurious signature mismatches, or (iii)~candidates that are
partially correct should receive credit for the cells they do get right.

We address these limitations with a \emph{tuple-level} consensus score.
Given the set of successfully executed candidates $\mathcal{C}^{+}$, we
first canonicalize each candidate's result into an order-independent
representation $\mathrm{frozenset}(\text{rows})$, then decompose it into
cell-level triples:
\begin{align}
\mathcal{T}(c)=\bigl\{(i,\mathit{col}_j,v_{ij}) \;\bigm|\;\,&
\text{row } i\text{, column } j\notag\\
&\text{of } \mathcal{E}(c,d)\bigr\},
\end{align}
where $\mathcal{E}(c,d)$ denotes the execution result of candidate $c$ on
database $d$.
The cross-pool frequency of a triple $t$ measures how many candidates agree
on that specific cell value:
\begin{equation}
f(t)=\bigl|\{c'\in\mathcal{C}^{+}\mid t\in\mathcal{T}(c')\}\bigr|.
\end{equation}
A candidate's consensus score then averages these frequencies over its own
non-null triples:
\begin{equation}
S_{\mathrm{maj}}(c)
=\frac{1}{|\mathcal{T}(c)|}
\!\sum_{t\in\mathcal{T}(c)}\!
f(t)\,\mathbf{1}[v(t)\neq\texttt{null}].
\end{equation}

\noindent \textbf{Relationship to whole-result frequency counting.}
When every candidate's result set is either identical or completely
disjoint, $S_{\mathrm{maj}}$ reduces to counting the number of candidates
that share the same signature (scaled by the result size), recovering
exact whole-result support counting as a special case.
The advantage appears precisely when result sets overlap partially: a
candidate that shares most cell values with the consensus receives a high
score even if its overall row set differs due to, e.g., a single extra
row or a different column alias.

\noindent \textbf{Empty-result exclusion and null down-weighting.}
Candidates returning zero rows have
$\mathcal{T}(c){=}\varnothing$, so $S_{\mathrm{maj}}(c){=}0$ by
convention.  This effectively excludes empty-result candidates from
the consensus, which is the primary filtering mechanism: such
candidates carry no execution evidence and should not compete with
candidates that produce substantive output.

For non-empty candidates, the indicator $\mathbf{1}[v(t){\neq}\texttt{null}]$
zeros out the frequency contribution of \texttt{null}-valued triples in the
numerator, while the denominator $|\mathcal{T}(c)|$ still counts all triples
including those with null values.
The effect is that \texttt{null} cells \emph{dilute} rather than inflate a
candidate's score: a result set dominated by nulls receives a lower
$S_{\mathrm{maj}}$ than one with the same number of non-null agreements.
This is deliberate rather than dismissive of nulls: \texttt{null} values
can reflect meaningful query logic (e.g., \texttt{LEFT JOIN} producing
\texttt{null} for unmatched rows), but because \texttt{null} cells are
pervasive across structurally different queries, counting their agreement
at full weight would inflate consensus scores without improving the
selector's discriminative power.

\subsection{Ablation Score Family: Freq, Majority Voting, Majority-Refine, and Merge}
\label{sec:selection-variants}

Table~\ref{tab:selection-ablation} in the main paper uses shorthand names for
several selectors derived from the same clean pool $\mathcal{C}^{+}$.
To avoid terminology ambiguity, we reserve \emph{freq} for exact-result
support counting, while \emph{majority voting} denotes the tuple-level
consensus score of Eq.~\ref{eq:smaj}, rather than exact-signature voting.

\noindent \textbf{Exact-result support / \emph{freq}.}
Let $\sigma(c)$ be the canonical result signature of candidate $c$.
We define the exact-result support count as
\begin{equation}
N_{\sigma}(c)
=\bigl|\{c'\in\mathcal{C}^{+}\mid \sigma(c')=\sigma(c)\}\bigr|.
\label{eq:freq-support-count}
\end{equation}
Its normalized frequency form is
\begin{equation}
S_{\mathrm{freq}}(c)=\frac{N_{\sigma}(c)}{|\mathcal{C}^{+}|}.
\label{eq:freq-score}
\end{equation}
Because the denominator is shared across candidates, ranking by
$S_{\mathrm{freq}}$ is identical to ranking by $N_{\sigma}$.
The \emph{freq} selector is therefore
\begin{equation}
y_{\mathrm{freq}}=\arg\max_{c\in\mathcal{C}^{+}} S_{\mathrm{freq}}(c).
\label{eq:freq-selector}
\end{equation}

\noindent \textbf{Tuple-level majority voting.}
Our \emph{majority voting} ablation keeps only the tuple-level consensus
signal:
\begin{equation}
y_{\mathrm{vote}}=\arg\max_{c\in\mathcal{C}^{+}} S_{\mathrm{maj}}(c).
\label{eq:maj-selector}
\end{equation}
This selector rewards candidates whose returned cells are repeatedly
supported across the pool, even when no other candidate matches their full
result signature exactly.

\noindent \textbf{Majority-refine.}
The \emph{majority-refine} statistic used in our implementation adds
exact-result support on top of tuple-level consensus:
\begin{align}
S_{\mathrm{ref}}(c)&=N_{\sigma}(c)+S_{\mathrm{maj}}(c),\notag\\
y_{\mathrm{ref}}&=\arg\max_{c\in\mathcal{C}^{+}} S_{\mathrm{ref}}(c).
\label{eq:ref-selector}
\end{align}
Using the unnormalized count $N_{\sigma}(c)$ keeps this term on the same raw
count scale as $S_{\mathrm{maj}}(c)$. In the released selector code, the gate
uses a closely related implementation-level quantity
\texttt{maj\_score\_refine} produced by weighted majority voting; conceptually
it serves the same role as a consensus-refine ambiguity check, even though its
exact tie-break behavior is defined by the implementation rather than by the
clean ablation formula above.

\noindent \textbf{RM-only and \emph{merge}.}
The pairwise preference ablation keeps only the position-debiased reward-model
score:
\begin{equation}
y_{\mathrm{rm}}=\arg\max_{c\in\mathcal{C}^{+}} S_{\mathrm{rm}}(c).
\label{eq:rm-selector}
\end{equation}
Our coarse backbone, denoted \emph{merge} in
Table~\ref{tab:selection-ablation}, fuses tuple-level consensus with
pairwise Elo preference:
\begin{align}
S_{\mathrm{merge}}(c)&=S_{\mathrm{maj}}(c)+S_{\mathrm{rm}}(c),\notag\\
y_{\mathrm{merge}}&=\arg\max_{c\in\mathcal{C}^{+}} S_{\mathrm{merge}}(c).
\label{eq:merge-selector}
\end{align}
The full system then applies the structural-support override
(\S\ref{sec:selection}) only when the confidence gate
fires.

\subsection{Position-Debiased Elo Derivation}
\label{sec:selection-elo}

We first partition the successful pool
$\mathcal{C}^{+}=\{c\mid\delta(c)=\texttt{clean}\}$
into result-signature equivalence classes:
\begin{align}
[\mathcal{C}^{+}]_\sigma
&= \bigl\{\{c\in\mathcal{C}^{+}\!\mid\!\sigma(c)=s\} \notag\\
&\qquad\bigm|\; s\in\mathrm{Im}(\sigma)\bigr\},
\label{eq:dedup-partition}
\end{align}
and pick one representative $\hat c_g$ per class.
For every representative pair $(\hat c_i,\hat c_j)$ we build a comparison
prompt containing the question, evidence, a schema summary restricted to
tables referenced by either candidate, and the first $k$ execution rows of
each.
Let $J_p(a,b)\in\{1,0,\bot\}$ denote the judge's verdict for candidate $a$
when it appears in position $p\in\{1,2\}$ of a comparison against $b$
($\bot$ flags a parse failure).
The position-debiased pairwise outcome for the ordered pair $(i,j)$ is
\begin{align}
\Delta_{i\succ j}
&= \bigl[J_1(\hat c_i,\hat c_j){-}J_2(\hat c_j,\hat c_i)\bigr]_{+}\notag\\
&\quad + \bigl[J_1(\hat c_j,\hat c_i){-}J_2(\hat c_i,\hat c_j)\bigr]_{-},
\end{align}
where $[\cdot]_{+}$ and $[\cdot]_{-}$ denote the positive and negative
parts. In practice we accumulate wins minus losses across all valid
comparisons, obtaining the cumulative score
\begin{align}
S_{\mathrm{rm}}^{\mathrm{dedup}}(\hat c_g)
= \!\!\sum_{(\hat c_g,\hat c_h)\in\mathcal{P}}\!\!
\bigl(&\mathbf{1}[J(\hat c_g){=}\mathrm{C}]\notag\\
- &\mathbf{1}[J(\hat c_h){=}\mathrm{C}]\bigr),
\end{align}
where $\mathrm{C}$ denotes the \texttt{Correct} verdict and $\mathcal{P}$
ranges over all $\binom{|[\mathcal{C}^{+}]_\sigma|}{2}$ valid
representative pairs.
The representative score is then broadcast back to class members:
\begin{equation}
S_{\mathrm{rm}}(c)=S_{\mathrm{rm}}^{\mathrm{dedup}}\!\bigl(\hat c_{g(c)}\bigr).
\end{equation}
This deduplication-then-broadcast scheme keeps the number of LLM judge calls
quadratic only in the number of distinct result signatures rather than in the
raw pool size.

\subsection{Pairwise Reward Model}
\label{sec:rm-training}

The pairwise reward model used to compute $S_{\mathrm{rm}}$ in
Eq.~\ref{eq:coarse} is a fine-tuned Qwen2.5-Coder-32B model that, given
two candidate SQLs together with their execution results, emits two
independent correctness labels wrapped in \texttt{<sql1\_judge>} and
\texttt{<sql2\_judge>} tags. Training has two stages: a supervised
warm-start checkpoint trained on the same pairwise judgment format, followed
by DAPO with a pairwise exact-match reward.

\noindent \textbf{Data.}
We synthesize pairwise examples from question-level candidate pools. For
each \bird question we collect multiple candidate SQLs under several
prompting strategies, execute each candidate on the target database to
obtain an individual correctness tag, and then materialize ordered pairs
$(\texttt{sql1},\texttt{sql2})$ with label tuple
$(y_1,y_2)\in\{\texttt{correct},\texttt{incorrect}\}^2$.
Table~\ref{tab:rm-data} summarizes the final train/validation split and its
realized label distribution. The four-class distribution is deliberate: an
effective pairwise judge must remain accurate when both candidates are
wrong (20.3\% of training pairs), not only when one is clearly better. The
equal counts of \texttt{correct,incorrect} and
\texttt{incorrect,correct} in the training split further indicate that both
orderings are retained rather than collapsed to a single example per
unordered pair.

\begin{table}[t]
\centering
\footnotesize
\setlength{\tabcolsep}{3pt}
\begin{tabular}{lrrrr}
\toprule
\textbf{Pair label} & \multicolumn{2}{c}{\textbf{Train}} & \multicolumn{2}{c}{\textbf{Val}} \\
\cmidrule(lr){2-3}\cmidrule(lr){4-5}
 & Count & \% & Count & \% \\
\midrule
\texttt{correct, correct}     & 7{,}298  &  9.2 &  788 & 75.6 \\
\texttt{correct, incorrect}   & 28{,}166 & 35.3 &  127 & 12.2 \\
\texttt{incorrect, correct}   & 28{,}166 & 35.3 &  127 & 12.2 \\
\texttt{incorrect, incorrect} & 16{,}170 & 20.3 &   -- &   -- \\
\midrule
Total                         & 79{,}800 &  --  & 1{,}042 & -- \\
\bottomrule
\end{tabular}
\caption{Pairwise RM training data, all sourced from \bird. Validation
contains no \texttt{incorrect, incorrect} pairs by construction, as both
candidates failing the execution gate indicate a degenerate pair that
carries no preference signal.}
\label{tab:rm-data}
\end{table}

\noindent \textbf{Supervised warm-start.}
We first fine-tune Qwen2.5-Coder-32B on the pairwise judgment format of
Listing with cross-entropy on the ground-truth
\texttt{<sql1\_judge>} and \texttt{<sql2\_judge>} tags as targets. We use
per-device batch size $1$, gradient accumulation $2$, learning rate
$1{\times}10^{-5}$ with cosine decay, and train for $3$ epochs. This stage
teaches the RM to emit a well-formed, parseable judgment before RL
introduces reward sparsity.

\noindent \textbf{Reinforcement learning with DAPO.}
We train the RM with DAPO~\citep{yu2026dapo} under a binary pairwise
exact-match reward. Let $j_1(y)$ and $j_2(y)$ denote the predicted
correctness tags parsed from \texttt{<sql1\_judge>} and
\texttt{<sql2\_judge>}, and $j_1^\star,j_2^\star$ the corresponding gold
tags. The reward is
\begin{equation}
R_{\mathrm{rm}}(y)
= \mathbf{1}[j_1(y){=}j_1^\star] \cdot \mathbf{1}[j_2(y){=}j_2^\star].
\label{eq:rm-reward}
\end{equation}
The indicator product means partial credit is not awarded: only
trajectories that correctly label \emph{both} SQLs receive reward $1$,
which forces the RM to reason about each SQL on its own merits rather
than anchoring on the obviously wrong one.
Table~\ref{tab:rm-hparams} lists the RM-specific hyper-parameters. The
DAPO objective, dynamic sampling, and threshold-gated entropy follow
\S\ref{sec:training-objective}.

\begin{table}[t]
\centering
\small
\begin{tabular}{lc}
\toprule
\textbf{Setting} & \textbf{Value} \\
\midrule
Rollout $n$                   & $16$ \\
Rollout temperature           & $1.0$ \\
Clip ratio low / high         & $0.2$ / $0.28$ \\
KL regularization             & disabled \\
Entropy target $H_{\mathrm{target}}$ & $0.15$ \\
Entropy coef.\ $\beta$        & $5{\times}10^{-3}$ \\
Max prompt / response         & $16{,}384$ / $1{,}024$ tokens \\
Overlong buffer               & $128$ tokens, factor $1.0$ \\
Train / gen / mini bsz        & $128$ / $384$ / $128$ \\
Learning rate                 & $1{\times}10^{-5}$ \\
LR warmup steps               & $10$ \\
Weight decay                  & $0.1$ \\
Gradient clip                 & $1.0$ \\
Total training steps          & $1{,}000$ \\
\bottomrule
\end{tabular}
\caption{Reward-model DAPO hyper-parameters. Longer prompts and shorter
responses (relative to the specialist) reflect the pairwise judgment
format: two SQLs with their execution results in, a pair of correctness
tags out.}
\label{tab:rm-hparams}
\end{table}

\subsection{Structural Signature Features}
\label{sec:selection-signature}

The AST-level structural signature $\kappa(c)$ used by the structural-support
reranker concatenates the following features extracted from the parsed SQL:
\begin{itemize}
\item $\mathsf{tables}(c)$: sorted tuple of base table names in FROM/JOIN.
\item $\mathsf{proj\_kinds}(c)$: tuple of projection kinds (e.g., \texttt{COUNT},
  \texttt{SUM}, \texttt{AVG}, plain column).
\item $\mathsf{proj\_count}(c)$: number of projected columns.
\item $\mathsf{has\_distinct}(c)$ and $\mathsf{has\_count\_distinct}(c)$:
  whether the query uses \texttt{DISTINCT} or
  \texttt{COUNT(DISTINCT}~$\cdot$\texttt{)}.
\item $\mathsf{has\_group\_by}(c)$ and $\mathsf{group\_by\_count}(c)$: whether
  the query uses \texttt{GROUP BY} and how many grouping expressions it uses.
\item $\mathsf{has\_order\_by}(c)$ and $\mathsf{order\_dirs}(c)$: whether the
  query uses \texttt{ORDER BY} and, if so, the direction tuple.
\item $\mathsf{limit\_bucket}(c)$: a coarse \texttt{LIMIT} bucket with three
  values: no limit, limit equals 1, and limit greater than 1.
\item $\mathsf{proj\_cols}(c)$: projected source-column tuple from SELECT.
\item $\mathsf{pred\_cols}(c)$: set of columns appearing in WHERE predicates.
\end{itemize}
The structural-support reranker of \S\ref{sec:selection} uses $\kappa(c)$
to measure \emph{cross-source consensus}: for each candidate class tied
in the coarse score, we count how many distinct source groups
(generator--route identities, with model-level fallback) contributed a
candidate with the same $\kappa$, and the class with the highest such count
wins the override if its support is strictly stronger than the baseline tied
winner.
Unlike a structural clustering, we do not build per-cluster evidence
summaries or invoke any LLM judge. The entire reranker is deterministic.

\section{Extended Discussion}
\label{sec:appendix-discussion}

This appendix expands the brief discussion stub at the end of \S\ref{sec:results}.

\subsection{The Coverage--Consensus Trade-off}

Our experiments surface a recurring pattern between specialist and generalist candidate sources.
Pools drawn from a single training distribution either reach a wider set of correct queries but cannot vote them out (high oracle, low majority), or they vote consistently but cannot reach the correct query in the first place (high majority, low oracle).
\system{} is effective on \bird{} not because it produces a strictly larger pool, but because pairing distributions sitting on opposite ends of this trade-off lifts both the ceiling and the voteability simultaneously (RQ3).
We expect this pattern to generalize to any setting where a domain-specialized RL-trained model can be paired with a broader LLM, beyond Text-to-SQL.

\subsection{Coarse Repair Can Pollute the Pool}

A second observation from RQ4 is more cautionary: handing failed candidates to a single LLM-based corrector without a structured per-type signal can be \emph{worse} than dropping those candidates entirely (\emph{coarse LLM repair} vs \emph{drop failures}).
This suggests that the value of execution feedback in pool-level pipelines lies in the structured diagnosis it carries, not in the LLM call it triggers: uniform LLM repair injects plausibly-clean noise that subsequent voting cannot filter out.
Practitioners adding self-correction to multi-candidate pipelines should preserve the failure type rather than collapse it.

\subsection{Efficiency--Specificity Trade-off in the Hybrid Selector}

The hybrid selector's design embodies a deliberate efficiency--specificity trade-off: most questions are resolved by lightweight result-level signals, and the deterministic structural-support reranker fires only on questions where these signals are tied.


\end{document}